\RequirePackage{fix-cm}

\documentclass[twocolumn]{svjour3}

\smartqed
  
\usepackage{amsmath}
\usepackage{hyperref}
\usepackage{cite}
\usepackage{doi}
\usepackage{multirow}
\usepackage{algorithmic}
\usepackage{arydshln}
\usepackage{afterpage}
\usepackage{subcaption}
\usepackage{graphicx}
\usepackage{color}
\usepackage{epstopdf}
\usepackage{bm}
\usepackage{textcomp}
\usepackage{microtype}
\usepackage{setspace}
\usepackage[toc,page]{appendix}
\newcommand{\DElscr}{DE$_{\text{lscr}}$}
\newcommand{\DEtl}{DE$_{\text{2P}}$}
\newcommand{\CALL}[2]{\texttt{{#1}(#2)}}
\newcommand{\IIF}[2]{
    \STATE\algorithmicif\ {#1}\ \algorithmicthen\ {#2} \algorithmicend\ \algorithmicif%
}
\newcommand{\IIFELSE}[3]{
    \STATE\algorithmicif\ {#1}\ \algorithmicthen\ {#2} \\ \algorithmicelse {~#3} \algorithmicend\ \algorithmicif%
}
\newcommand{\IDOWHILE}[2]{
    \STATE\algorithmicdo\ {#1}\ \algorithmicwhile\ {#2} \algorithmicend~\algorithmicdo%
}

\newcommand{\PROCEDURE}[1]{%
	\renewcommand{\algorithmicwhile}{\textbf{procedure}}
	\renewcommand{\algorithmicdo}{ }
	\WHILE{#1}
	\renewcommand{\algorithmicwhile}{\textbf{while}}
	\renewcommand{\algorithmicdo}{\textbf{do}}	
}
\newcommand{\ENDPROCEDURE}{\ENDWHILE}
\definecolor{Gray}{rgb}{0.9,0.9,0.9}
\newcommand{\markg}[1]{\setlength{\fboxsep}{2pt}{\colorbox{Gray}{#1}}}

\begin{document}
\title{Two-phase protein folding optimization on a three-dimensional AB off-lattice model}
\titlerunning{Two-phase protein folding optimization on a 3D AB off-lattice model}
\author{Borko Bo\v{s}kovi\'{c} \and Janez Brest}
\institute{B. Bo\v{s}kovi\'{c} \and J. Brest \at
    Faculty of Electrical Engineering and Computer Science,\\ University of Maribor, SI-2000 Maribor, Slovenia \\
    \email{borko.boskovic@um.si, janez.brest@um.si}
}
\renewcommand{\labelitemi}{$\bullet$}

\date{\today}

\maketitle

\begin{abstract}
This paper presents a two-phase protein folding optimization on a three-dimen-sional AB off-lattice model. 
The first phase is responsible for forming conformations with a good hydrophobic core or a set of compact hydrophobic
amino acid positions. These conformations are forwarded to the second phase, where an accurate search is performed
with the aim of locating conformations with the best energy value. The optimization process switches between these two 
phases until the stopping condition is satisfied. An auxiliary fitness function was designed for the first phase, while 
the original fitness function is used in the second phase. The auxiliary fitness function includes an expression about 
the quality of the hydrophobic core. This expression is crucial for leading the search process to the promising 
solutions that have a good hydrophobic core and, consequently, improves the efficiency of the whole optimization process. 
Our differential evolution algorithm was used for demonstrating the efficiency of two-phase optimization.
It was analyzed on well-known amino acid sequences that are used frequently in the literature. The obtained experimental
results show that the employed two-phase optimization improves the efficiency of our algorithm significantly and that the
proposed algorithm is superior to other state-of-the-art algorithms. 

\keywords{Protein folding optimization \and AB off-lattice model \and Differential evolution \and Two-phase optimization}
\end{abstract}

\section{Introduction}
\label{sec:intro}
Proteins are fundamental components of cells in all living organisms. They perform many tasks,
such as catalyzing certain processes and chemical reactions, transporting molecules to and from the cell,
delivering messages, sensing signals, and other things which are essential for the preservation of life~\cite{Jana2018}.
Proteins are formed from one or more amino acid chains joined together. The amino acid chain must fold into a
specific three-dimensional native structure before it can perform its biological function(s)~\cite{petsko2004}.
An incorrectly folded structure may lead to many human diseases, such as Alzheimer's disease, cancer, and cystic
fibrosis. Therefore, the problem of how to predict the native structure of a protein from its amino acid
sequence is one of the more important challenges of this century~\cite{Kennedy05} and, because of its nature,
it attracts scientists from different fields, such as Physics, Chemistry, Biology, Mathematics, and Computer
Science. 

Scientists are trying to solve the protein structure prediction problem with experimental and computational methods.
The experimental methods, such as X-ray crystallography and nuclear magnetic resonance, are very time consuming and  
expensive. In order to mitigate these disadvantages of experimental methods, scientists are trying to develop 
computational methods. Template based methods use information about related or similar sequences. In contrast to these
methods, \emph{ab-initio} methods predict the native three-dimensional structure of an amino acid chain from its 
sequence, and, to do this, they do not require any additional information about related sequences. They predict the 
three-dimensional structure from scratch. These methods are not only important because they are an alternative to 
experimental methods, but also because they can help to understand the mechanism of how proteins fold in nature. 
Therefore, inside \emph{ab-initio} methods, the Protein Folding Optimization (PFO) represents a computational problem of 
how to simulate the protein folding process and to find a native structure. Improving PFO will lead to the improvement of 
prediction methods and, consequently, this could reduce the gap between the number of known protein sequences and known 
protein structures.

Using \emph{ab-initio} methods, it is possible to predict the native structure of relatively small proteins. The reasons
for that are an expensive evaluation of conformation, and the huge and multimodal search space. In order to reduce the 
time complexity of evaluations and to reduce the spatial degrees of freedom, simplified protein models were designed, such
as an HP model~\cite{Hart2005,BoskovicHP16} within different lattices, and an AB off-lattice model~\cite{Stillinger93}.
The main goals of these models are development, testing, and comparison of different methods. Within this 
paper, the simplified three-dimensional AB off-lattice model was used to demonstrate the efficiency of two-phase 
optimization by using the Differential Evolution (DE) algorithm.

It has been shown that PFO has a highly rugged landscape structure, containing many local optima and needle-like
funnels~\cite{Jana17a}. In order to explore this search space effectively, we have already proposed a 
DE algorithm~\cite{BoskovicAB16,BoskovicAB18} that, in contrast to all previous methods, follows only
one attractor. The DE algorithm was selected because of its simplicity and efficiency, and because it was 
used successfully in various optimization problems~\cite{Karol19,Das16,Das11,DelSer19}, such as an animated trees 
reconstruction~\cite{Zamuda14}, a post hoc analysis of sport performance~\cite{Fister18}, and parametric design and 
optimization of magnetic gears~\cite{Wang19}. The temporal locality~\cite{Wong12}, self-adaptive 
mechanism~\cite{Brest06} of the main control parameters, local search, and component reinitialization were used 
additionally to improve the efficiency of our algorithm. The DE algorithm, with all its listed mechanisms, was capable of 
obtaining significantly better results than other state-of-the-art algorithms, and it obtained a success ratio of 100\% 
for sequences up to 18 monomers.

In this paper, we propose a new two-phase optimization DE algorithm. In contrast to bilevel 
optimization, where an optimization problem contains another optimization problem as a constraint~\cite{Sinha18,Kumar19},
our approach uses two optimization phases. The auxiliary fitness function is designed for the first phase, and the original
fitness function is used in the second phase. The auxiliary fitness function contains an expression that allows the algorithm
to locate solutions with a good hydrophobic core easily. The hydrophobic core represents a set of positions of the hydrophobic
amino acids. The motivation for this approach is taken from nature, where the hydrophobic amino acids hide from water and form
the hydrophobic core, while hydrophilic amino acids move to the surface to be in contact with the water molecules. To
simulate this process from nature, the first phase is responsible for locating solutions with a good hydrophobic core quickly,
while the second phase is responsible for final optimization. The optimization process continues with reinitializations, and
alternates between these two phases until the stopping condition is satisfied. Note that reinitializations are performed after
the second phase. A component reinitialization was used to change only a few components in individuals, with the
purpose to get out of the corresponding valley or funnel in the search space, and to locate better solutions, while a random
reinitialization generates all components in individuals randomly, and guides the search process to unexplored search space
regions. If we perform reinitializations after the first phase, the algorithm will locate fewer solutions with a
good hydrophobic core, and, therefore, the algorithm may not be able to find the global optimum at all. We called the
proposed algorithm \DEtl\ (Two-phases Differential Evolution), and it was tested on two sets of amino acid sequences that 
were used frequently in the literature.
The first set includes 18 real peptide sequences, and the second set includes 5 well-known artificial Fibonacci sequences.
Experimental results show that the proposed two-phase optimization improves the efficiency of the algorithm, and it is
superior to other state-of-the-art algorithms. Our algorithm is now capable of reaching the best-known conformations with
a success rate of 100\% for sequences up to 25 monomers within the budget of $10^{11}$ solution evaluations. The new best-known
solutions were reached by the algorithm for all sequences with a length of 29 or more monomers. Based on these observations,
the main contributions of this paper are:
\begin{itemize}
  \item The two-phase optimization.
  \item The auxiliary fitness function.
  \item The frontiers of finding the best-known solutions with a success rate of 100\% are pushed to the sequences with 
up to 25 monomers.
  \item The new best-known conformations for all sequences with 29 or more monomers.
\end{itemize}

\noindent
The remainder of the paper is organized as follows. Related work and the three-dimensional AB off-lattice model
are described in Sections~\ref{sec:related} and~\ref{sec:model}. The two-phase optimization DE
algorithm and auxiliary fitness function are given in Section~\ref{sec:method}. The description of the experiments and 
obtained results are presented in Section~\ref{sec:experiments}. Section~\ref{sec:conclusions} 
concludes this paper.

\section{Related work}
\label{sec:related}
Over the years, different types of metaheuristic optimization algorithms have been applied successfully to the PFO on
the AB off-lattice model. A brief overview of the existing algorithms is provided within this Section.
The information about hydrophobic cores was used within different approaches for protein structure prediction. 
A brief description of these approaches is also included in this Section.

\subsection{Metaheuristic optimization algorithms}
Evolutionary algorithms have been quite successful in solving PFO.
An ecology inspired algorithm for PFO is presented in~\cite{Parpinelli13}. A key concept of this algorithm is the 
definition of habitats. These habitats, or clusters, are determined by using a hierarchical clustering algorithm.
For example, in a multimodal optimization problem, each peak can become a promising habitat for some populations. Two 
categories of ecological relationships can be defined, according to the defined habitats, intra-habitats' relationships 
that occur between populations inside each habitat, and inter-habitats' relationships that occur between habitats. The 
intra-habitats' relationships are responsible for intensifying the search, and the inter-habitats' relationships are 
responsible for diversifying the search.

Paper~\cite{Kalegari13} presents the basic and adaptive versions of the DE algorithm with parallel architecture 
(master-slave). With this architecture, the computational load is divided and the overall performance is improved.
An explosion and mirror mutation operators were also included into DE. The explosion is a mechanism that reinitializes
the population when the stagnation has occurred, and, thus, it is responsible for preventing premature convergence.
The second mechanism, the mirror mutation, was designed to perform a local search by using mirror angles within the 
sequence.

In paper~\cite{Sar14}, the authors have analyzed six variants of Genetic Algorithms (GAs). Three variants were designed, and 
each of them includes one of the following selection mechanisms: Rank selection, elitist selection, and tournament 
selection. All of these variants are combined with single and double point crossover. The GA with the elitist selection
and two-point crossover outperforms other variants.

A Biogeography-Based Optimization (BBO) is also applied to PFO~\cite{Fan14}. This algorithm is based on the
definition of habitats. Each habitat has its amount of species, and different habitats usually have different amounts
of species. Within the algorithm, Habitat Suitability Index (HSI) is used to measure the quality of the habitat.
Habitats with high HSI are suitable for survival. Thus, these habitats have low immigration rates and high emigration 
rates. On the contrary, habitats with low HSI have high immigration rates and low emigration rates. Additionally, BBO 
includes a mutation operator to avoid premature convergence, and elitism to avoid the degeneration phenomena.
The improved BBO contains an improved migration process. In the  migration process, a feature from a habitat is replaced by 
another feature from a different habitat. In the improved version, different features were selected from different 
habitats according to their emigration rates, and their values with weights determine the features of the habitat. This 
algorithm was compared with the standard BBO and DE. The results show that the improved BBO outperforms all competitors.

It has been shown that the PFO has a highly rugged landscape structure containing many local optima and needle-like
funnels~\cite{Jana17a}, and, therefore, the algorithms that follow more attractors simultaneously are ineffective. In 
our previous work~\cite{BoskovicAB16}, to overcome this weakness, we proposed the DE algorithm that uses the 
\texttt{best/1/bin} strategy. With this strategy, our algorithm follows only one attractor. The temporal locality 
mechanism \cite{Wong12} and self-adaptive mechanism \cite{Brest06} of the main control parameters were used additionally 
to speed up the convergence speed. Random reinitialization was used when the algorithm was trapped in a local optimum.
This algorithm was improved in \cite{BoskovicAB18} with two new mechanisms. A local search is used to improve 
convergence speed, and to reduce the runtime complexity of the energy calculation. For this purpose, a local movement is 
introduced within the local search. The designed evolutionary algorithm has fast convergence speed and,
therefore, when it is trapped into the local optimum or a relatively good solution is located, it is hard to locate a better
similar solution. The similar solution may differ from the good solution in only a few components. A component 
reinitialization method is designed to mitigate this problem. It changes only a few components in individuals, 
with the purpose to get out of the corresponding valley in the search space and to locate better solutions.
The obtained results of this algorithm show that it is superior to the algorithms from the literature, and significantly
lower energy values were obtained for longer sequences. 

Swarm Intelligence algorithms also showed good results for PFO. The authors in~\cite{Parpinelli14} tested the 
standard versions of the following algorithms: Particle swarm optimization, artificial bee colony, gravitational search 
algorithm, and the bat algorithm. This test showed that the particle swarm optimization algorithm obtained
the overall best balance between quality of solutions and the processing time.

To improve the Artificial Bee Colony (ABC) algorithm convergence performance, an internal feedback strategy based ABC is 
proposed in \cite{LI14}. In this strategy, internal states are used fully in each iteration, to guide the subsequent 
searching process. In~\cite{Wang13}, a chaotic ABC algorithm was introduced. This algorithm combines the artificial bee 
colony and the chaotic search algorithm to avoid the premature convergence. If the algorithm was trapped into 
the local optimum, it uses a chaotic search algorithm to prevent stagnation. A balance-evolution artificial bee colony 
algorithm was presented in \cite{Li15f,Li15}. During the optimization process, this algorithm uses convergence 
information to manipulate adaptively between the local and global searches. For a detailed description of the ABC algorithm,
we refer readers to~\cite{Rajasekhar17}.

Researches combined two or more algorithms in order to develop hybrid algorithms that can obtain better results in 
comparison with the original algorithms. In~\cite{Lin14}, the authors combined simulated annealing and the tabu search 
algorithm. This algorithm was improved additionally with a local adjust strategy that improves the accuracy and speed of 
searching.

The algorithm that combines the particle swarm optimization, genetic algorithm, and tabu search was presented 
in~\cite{Zhou14a}. Within this algorithm, the particle swarm optimization is used to generate an initial solution
that is not too random, and the factor of stochastic disturbance is adopted to improve the ability of the global
search. The genetic algorithm was used to generate local optima in order to speed up the convergence of the
algorithm, while the tabu search is used with a mutation operator to locate the global optimum.

An improved stochastic fractal search algorithm was applied to the AB off-lattice model in~\cite{Zhou18}. In order to avoid 
the algorithm becoming trapped into the local optimum, L{\'e}vy flight and internal feedback information were
incorporated into the algorithm. The algorithm consists of diffusion and an update process. The L{\'e}vy flight was used
in the diffusion process to generate some new particles around each population particle. In the update process, the best particle 
generated from the diffusion process is used to generate new particles. To prevent stagnation within a local optimum, 
the internal feedback information is incorporated into the algorithm. This information is used to trigger the mechanism 
that generates new particles according to two randomly selected particles from the population.

The authors in~\cite{Hribar18} have shown that the DE algorithm converges to better solutions when 
the initial population is created by using trained neural networks. The neural networks were trained successfully using 
the reinforcement learning method, by knowing only the fitness function of the class of optimization problems.

An improved harmony search algorithm was presented in~\cite{Jana2018,Jana17b}. In this algorithm, the basic harmony 
search algorithm was combined with the dimensional mean based perturbation strategy. This strategy allows the algorithm to 
avoid premature convergence, and enhance the capability of jumping out from the local optima. 

A multi-agent simulated annealing algorithm with parallel adaptive multiple sampling was proposed in~\cite{Lin18}.
A parallel elitist sampling strategy was used to overcome the inherent serialization of the original simulated 
annealing algorithm. This strategy additionally provides benefit information, that is helpful for the convergence. An 
adaptive neighborhood search and a parallel multiple move mechanism were also used inside the algorithm to improve 
the algorithm's efficiency. In this work, the following methods were analyzed for generating candidate solutions:
Simulated annealing, a mutation from the DE algorithm, and the velocity and position update 
from the particle swarm optimization.

Although powerful optimization algorithms have been introduced for PFO, researchers are also focused on the 
time-consuming optimization problems. For solving such a problem for PFO, the authors in~\cite{Rakhshani19} introduced a new
version of DE which uses computationally cheap surrogate models and gene expression programming. The purpose of the 
incorporated gene expression programming is to generate a diversified set of configurations, while the purpose of the 
surrogate model is to help DE to find the best set of configurations. Additionally, a covariance matrix adaptation 
evolution strategy was also adopted, to explore the search space more efficiently. This algorithm is called SGDE, and it 
outperforms all state-of-the-art algorithms according to the number of function evaluations. Its efficiency was also 
demonstrated in terms of runtime on the adopted all-atom model which represents time-consuming PFO.

\newcommand{\centerhfill}[1][\quad]{\hspace{\stretch{0.5}}#1\hspace{\stretch{0.5}}}
\begin{figure*}
 \centering
 \begin{subfigure}[t]{0.49\textwidth}
 \centering
\includegraphics[scale=0.24]{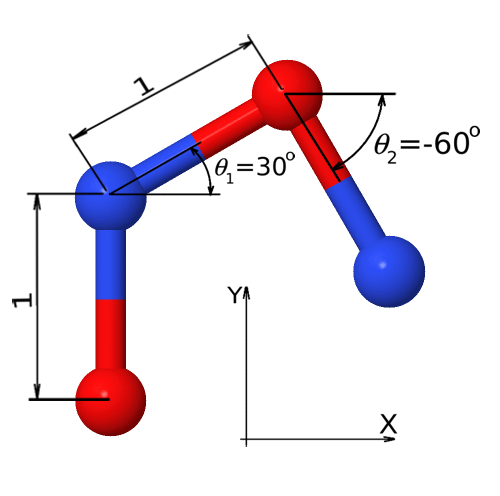}
\caption{}
 \end{subfigure}
 \begin{subfigure}[t]{0.49\textwidth}\centering
 \centering
\includegraphics[scale=0.24]{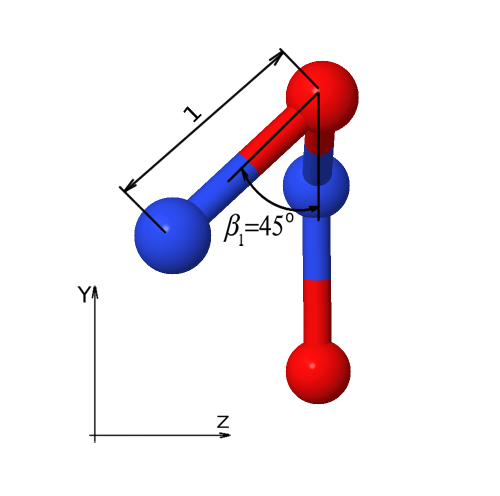}
\caption{}
 \end{subfigure}
 \caption{A schematic diagram of a sequence ABAB. (a) Projection of a structure with $\theta_1=30$, $\theta_2=-60$ and 
    $\beta_1=0$ onto the XY-plane. (b) Projection of a structure with $\theta_1=30$, $\theta_2=-60$ and $\beta_1=45$ onto 
    the ZY-plane.}
 \label{fig:model}
\end{figure*}

\subsection{Hydrophobic core}
Information about a hydrophobic core, or a set of positions of the hydrophobic amino acids, is very useful for structure
prediction in different methods. The authors in~\cite{Backofen06} presented a constraint-based method. The key concept of 
this method is the ability to compute maximally compact hydrophobic cores. Information about hydrophobic cores was also 
used within stochastic algorithms for PFO. In~\cite{Rashid16a,Rashid16b,Rashid13} a macro-mutation operator is incorporated into 
the genetic algorithm and applied to the three-dimensional face-centered cubic lattice. This operator compresses the 
conformation, and forms the hydrophobic-core quickly. The obtained results show that the macro-mutation operator 
improves the efficiency of the algorithm significantly.

\section{Three-dimensional AB off-lattice model}
\label{sec:model} 

A chain of amino acids can be represented with a unique amino acid sequence. From the amino acid sequence, it is possible
to generate different conformations, which is also dependent on the model used. There are different models,
such as a full atom, coarse-grained~\cite{Kmiecik16}, AB off-lattice~\cite{Stillinger93} and HP~\cite{Hart2005,BoskovicHP16}.
The full atom model is the most accurate and complex, and, as such, is computationally expensive. Although it is 
computationally expensive it was also used in the optimization process of the DE algorithm that was adapted for such a
problem~\cite{Rakhshani19}. Different coarse-grained models were designed to mitigate the computational complexity of the full
atom model~\cite{Kmiecik16}. The remaining two models, AB off-lattice and HP, are even more simplified and, as such, they are
not very computationally expensive. The main purpose of these models is an analysis of different optimization algorithms. 
The three-dimensional AB off-lattice model is used in our paper, to demonstrate the efficiency of the proposed algorithm.
Instead of 20 standard amino acids, this model uses only two different types of amino acids: $A$ -- hydrophobic and $B$ --
hydrophilic. Thus, an amino acid sequence is represented as a string $\vec{s}=\{s_1, s_2, ..., s_L\}$, $s_i \in \{A, B\}$,
where $A$ represents a hydrophobic, $B$ a hydrophilic amino acid, and $L$ the length of the sequence. The solution, or
three-dimensional structure of an AB sequence, is defined by bond angles $\vec{\theta}=\{\theta_1,\theta_2, ..., \theta_{L-2}\}$,
torsional angles $\vec{\beta}=\{\beta_1, \beta_2, ..., \beta_{L-3}\}$, and the unit-length chemical bond between two consecutive 
amino acids (see Fig.~\ref{fig:model}). The quality of the solution determines the energy value $E_{o}$, which is calculated
using a simple trigonometric form of backbone bend potentials $E_{\text{bb}}(\vec{\theta})$ and a species-dependent
Lennard-Jones 12,6 form of non-bonded interactions $E_{\text{lj}}(\vec{s},\vec{\theta},\vec{\beta})$, as shown in the
following equation~\cite{Stillinger93}:

\begin{footnotesize}
\begin{eqnarray}
  \footnotesize
  \label{eq:energy}
  E_{\text{o}}(\vec{s},\vec{\theta},\vec{\beta}) &=& E_{\text{bb}}(\vec{\theta})+E_{\text{lj}}(\vec{s},\vec{\theta},
  \vec{\beta}) \nonumber \\
  E_{\text{\tiny bb}}(\vec{\theta}) &=& \frac{1}{4}\displaystyle\sum_{i=1}^{L-2}[1-cos(\theta_i)]\\
  E_{\text{\tiny lj}}(\vec{s},\vec{\theta},\vec{\beta}) &=& 4\displaystyle\sum_{i=1}^{L-2}\displaystyle
  \sum_{\text{{\it j}={\it i}+}2}^{L}[d(\vec{p}_i,\vec{p}_j)^{\text{--}12} \text{--} c(s_i,s_j)\cdot 
d(\vec{p}_i,\vec{p}_j)^{\text{--}6}] \nonumber
\end{eqnarray}
\end{footnotesize}

\noindent
where $\vec{p}_i$ and $\vec{p}_j$ represent the position of the amino acid within the three-dimensional space.
These positions are determined, as shown in Fig.~\ref{fig:model} and by the following equation:

\begin{footnotesize}
\begin{eqnarray}
 \label{eq:positions}
 \vec{p}_i=\begin{cases}
	\{0,0,0\}&\text{if}~i=1,\\
	\{0,1,0\}&\text{if}~i=2,\\
	\{cos(\theta_1),1+sin(\theta_1),0\}&\text{if}~i=3,\\
	\{x_{i-1}+cos(\theta_{i-2})\cdot cos(\beta_{i-3}),\\ 
	  ~~y_{i-1}+sin(\theta_{i-2})\cdot cos(\beta_{i-3}),&\text{if 4}\le i\le L.\\
	 ~~z_{i-1}+sin(\beta_{i-3})\}\\
	\end{cases}
\end{eqnarray}
\end{footnotesize}

\noindent
In Eq.~(\ref{eq:energy}) $d(\vec{p}_i,\vec{p}_j)$ denotes the Euclidean distance between positions $\vec{p}_i$ and 
$\vec{p}_j$, while $c(s_i,s_j)$ determines the attractive, weak attractive or weak repulsive non-bonded interaction for 
the pair $s_i$ and $s_j$, as shown in the following equation:

\begin{footnotesize}
\begin{eqnarray}
 c(s_i,s_j) = \begin{cases}
                1    & \text{if }s_i = A \text{ and } s_j = A,\\ 
                0.5 & \text{if }s_i = B \text{ and } s_j = B,\\
                -0.5  & \text{if }s_i \neq s_j.\\
               \end{cases}
               \nonumber
\end{eqnarray}
\end{footnotesize}

\noindent
The objective of PFO within the context of an AB off-lattice model is to simulate the folding process, and to find the 
angles' vector or conformation that minimizes the free-energy value:
$\{\vec{\theta}^*,\vec{\beta}^*\} = \operatorname*{\mathrm{arg\,min}} E_{o}(\vec{s},\vec{\theta},\vec{\beta})$.
The described  model takes into account the hydrophobic interactions which represent the main driving forces of a
protein structure formation and, as such, still imitates its main features realistically~\cite{Huang06}. Therefore, 
although this model is incomplete, it allows the development, testing, and comparison of various search algorithms. 

\section{Method}
\label{sec:method}
\begin{figure}[t!]
\centering
\includegraphics[scale=0.40]{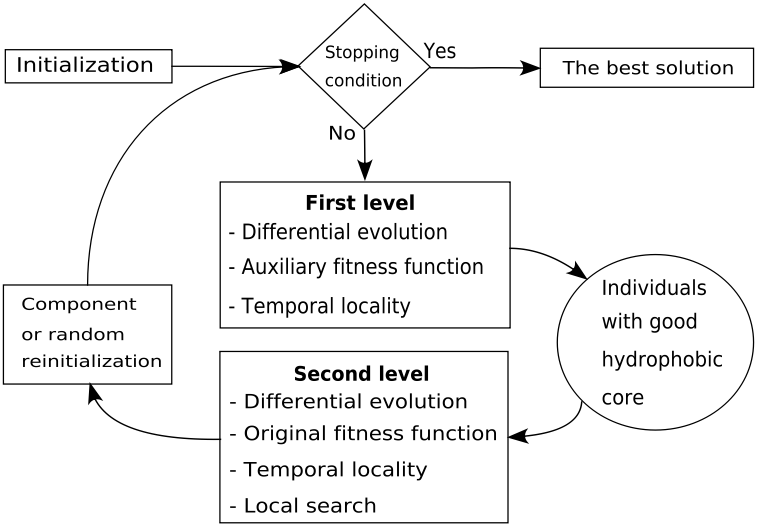}
\caption{Two-phase optimization process of the proposed DE algorithm.}
\label{fig:DEtl}
\end{figure}

In order to include knowledge about the hydrophobic core to our algorithm, we have developed the two-phase optimization 
by using the DE algorithm. The optimization process is alternated between two phases until the 
stopping condition of the optimization process is satisfied, as shown in Fig.~\ref{fig:DEtl}. The auxiliary fitness function
is used in the first phase, which is responsible for forming conformations with a good hydrophobic core. When the stopping
condition of the first phase is satisfied, the obtained population is forwarded to the second phase. The original fitness
function is used in the second phase to locate solutions with the lowest energy value. When the second optimization phase
is finished, the reinitialization is performed, and the optimization process continues with the first phase.

The main idea of the auxiliary fitness function is to allow the algorithm to form good hydrophobic-cores easily. For
this purpose, it contains three expressions, as shown in Eq. (\ref{eq:aux}).

\begin{footnotesize}
\begin{eqnarray}
  \label{eq:aux}
  E_{\text{x}}(\vec{s},\vec{\theta},\vec{\beta}) &=& E_{o}(\vec{s},\vec{\theta},\vec{\beta}) + 
  E_{\text{hc}}(\vec{s},\vec{\theta},\vec{\beta}) \nonumber
  + \lambda \nonumber \\
  E_{\text{hc}}(\vec{s},\vec{\theta},\vec{\beta}) & = & \sum_{i=1}^{L} d(\vec{p}_i,\vec{c}) \cdot h(s_i) \\
  h(s_i) & = & \begin{cases}
                1 & \text{if }s_i = A,\\ 
                0 & otherwise \nonumber
               \end{cases}\\
  \vec{c} & = & \frac{\sum_{i}^{N}{\vec{p}_i\cdot h(s_i)}}{N_A}\nonumber;~~~ N_A = \sum_i^N h(s_i)\nonumber
\end{eqnarray}
\end{footnotesize}

\noindent
The first expression, $E_{o}$, represents the original fitness function, the second expression, $E_{hc}$, determines the 
quality of the hydrophobic core, while the third expression is a constant $\lambda$ that separates the fitness values between
the first and second optimization phases. The quality of the hydrophobic core determines the sum of the Euclidean distances
between all hydrophobic amino acids and their centroid $\vec{c}$. The value $\lambda$ ensures that the
energy value $E_{x}$ in the first phase is always worse in comparison with the energy value $E_o$ in the second
optimization phase. In this work, the value of $\lambda$ was set to 1,000. This value is large enough to make a gap between
the fitness values of both optimization phases. This can be seen in the convergence graph (Fig. \ref{fig:con}) where the red line represents
the fitness values of the first phase, while the green line represents the fitness values of the second phase. The introduced gap
between phase values allows the algorithm to update the global and local best vectors correctly without any additional mechanism.

\subsection{Proposed algorithm}
The proposed algorithm is based on our previous work~\cite{BoskovicAB18}, and it was improved in such a way that
the optimization process is divided into two phases. This algorithm is described in this Section, and it is shown in Figs.~\ref{alg:detl} 
and~\ref{alg:reinitialization}. The lines that include two-phase optimization into our algorithm are highlighted with a gray
background.

\begin{figure}[!t]
\tiny
\begin{spacing}{1.2}
\begin{algorithmic}[1]
\algsetup{linenosize=\tiny}
\PROCEDURE{\CALL{\DEtl}{$\vec{s},\mathit{Np}$}}	
\STATE {Initialize a population $P$ \label{alg:init} \\ 
	\markg{$\mathit{firstPhase} = \text{\bf true}$}\\
	$\{ \vec{x}_{i}, F_{i}=0.5, \mathit{Cr}_{i}=0.9, e_i=$\markg{\CALL{$E_{x}$}{$\vec{s},\vec{x_i}$}}$\} \in P$ \\
	$ x_{i,j} = -\pi + 2 \cdot \pi \cdot \mathit{rand}_{[0,1]}$ \\
	$ i = 1, 2, ..., \mathit{Np}; j = 1, 2, ..., D; D = 2 \cdot $ \CALL{$\mathit{length}$}{$\vec{s}$} - 5}\\
	$ \{\vec{x}_{b},e_{b}\} = \{\vec{x}_{b}^{l},e_{b}^{l}\} = \{\vec{x}_{b}^{p},e_{b}^{p}\} = $ \CALL{BEST}{$P$}
\WHILE {stopping criteria is not met} \label{alg:gen}
	\FOR{$i=1$ \TO $\mathit{Np}$}
		\IIFELSE{$\mathit{rand}_{[0,1]} < 0.1 $} {$F = 0.1 + 0.9 \cdot \mathit{rand}_{[0,1]} $}{$F = F_{i}$} 
\label{alg:jde:a}
		\IIFELSE{$\mathit{rand}_{[0,1]} < 0.1 $}{$\mathit{Cr} = \mathit{rand}_{[0,1]}$}{$\mathit{Cr} = \mathit{Cr}_{i}$} 
\label{alg:jde:b}
		\IDOWHILE{$r_{_1}\text{=}\mathit{rand}_{\{1,\mathit{Np}\}}$}{$r_{_1}\text{=}i$} \label{alg:str:a}
		\IDOWHILE{$r_{_2}\text{=}\mathit{rand}_{\{1,\mathit{Np}\}}$}{$r_{_2}\text{=}i~\text{\bf 
or}~r_{_2}\text{=}r_{_1}$}
		\STATE{$j_{\mathit{rand}} = \mathit{rand}_{\{1,\mathit{D}\}}$}
		\FOR{$j=1$ \TO $D$}
 			\IF{$\mathit{rand}_{[0,1]} < \mathit{Cr}~\text{\bf or }~j = j_{\mathit{rand}}$}
 				\STATE{$ u_{j} = x_{b,j}+F\cdot(x_{r_1,j}-x_{r_2,j} ) $}
 				\IIF{$u_{j}\le\text{-}\pi$}{$u_{j} \text{\,=\,} 2\cdot\pi+u_{j}$}
 				\IIF{$u_{j}>\pi$}{$u_{j} \text{\,=\,} 2\cdot(\text{-}\pi)+u_{j}$}
 			\ELSE
				\STATE{$u_{j} = x_{i,j}$}
 			\ENDIF
		\ENDFOR \label{alg:str:b}
		\IF{\markg{$\mathit{firstPhase}$}} \label{alg:if:aux1}
			  \STATE{\markg{$e_u=$\CALL{$E_{x}$}{$\vec{s},\vec{u}$} // Auxiliary fitness function}} \label{alg:aux1}
		\ELSE
			  \STATE{\markg{$e_u=$\CALL{$E_{o}$}{$\vec{s},\vec{u}$} // Original fitness function}} \label{alg:eval1}
		\ENDIF \label{alg:jde:end}
		\IF{$e_u \le e_i$} \label{alg:tl:a}
			\STATE{// Temporal locality} 
			\FOR{$j=1$ \TO $D$}
				\STATE{$u^*_{j} = x_{b,j} + 0.5 \cdot (u_{j} - x_{i,j}) $}
 				\IIF{$u^*_{j}\le\text{-}\pi$}{$u^*_{j} \text{\,=\,} 2\cdot\pi+u^*_{j}$}
 				\IIF{$u^*_{j}>\pi$}{$u^*_{j} \text{\,=\,} 2\cdot(\text{-}\pi)+u^*_{j}$}
			\ENDFOR \label{alg:tl:b}
			\IF{\markg{$\mathit{firstPhase}$}} \label{alg:if:aux2}
				\STATE{\markg{$e^*_u=$\CALL{$E_{x}$}{$\vec{s},\vec{u}^*$} // Auxiliary fitness function}} 
\label{alg:aux2}
			\ELSE
				\STATE{\markg{$e^*_u=$\CALL{$E_{o}$}{$\vec{s},\vec{u}^*$} // Original fitness function}} 
\label{alg:eval2}
			\ENDIF
			\IF{$e^*_{u} \le e_u$}
				\STATE{$\{\vec{x_i},F_i,\mathit{Cr_i},e_i\}=\{u^*,F,\mathit{Cr},e^*_{u}\}$} \label{alg:s1}
			\ELSE
				\STATE{$\{\vec{x_i},F_i,\mathit{Cr_i},e_i\}=\{u,F,\mathit{Cr},e_{u}\}$} \label{alg:s2}
			\ENDIF
			\IF{\markg{{\bf not} $\mathit{firstPhase}$}} \label{alg:if:local}
				\STATE {// Local Search}
				\FOR{$n=2$ \TO $L-1$}
					\STATE {$\theta_{n\text{-}1} = \mathit{rand}_{[0,1]}\cdot ( x^{p}_{b,n\text{-}1} - x_{i,n\text{-}1} 
)$}
					\STATE {$\beta_{n\text{-}2} = \mathit{rand}_{[0,1]}\cdot ( x^{p}_{b,n+(L\text{-}4)} - 
x_{i,n+(L\text{-}4)})$}
					\STATE {\{${\vec{v}, e_v}\} = $ 
\CALL{LOCAL\_MOVEMENT}{$\vec{x}^{p}_{b},n,\theta_{n\text{-}1},\beta_{n\text{-}2}$}} \label{alg:lm}
					\IIF{{$e_{v}\le e_{b}$}}{$ \{\vec{x}^{p}_{b},e^{p}_{i}\}=\{\vec{v},e_{v}\}$}
				\ENDFOR
			\ENDIF
		\ENDIF \label{alg:impr:b}
	\ENDFOR
	\STATE {$ \{\vec{x}_{b}^{p},e_{b}^{p}\} = $ \CALL{BEST}{$P$}}
	\IIF{$e_{b}^{p} \le e_{b}$}{$ \{\vec{x}_{b},e_{b}\} = \{\vec{x}_{b}^{p},e_{b}^{p}\}$ }
	\STATE 
\markg{\CALL{REINITIALIZATION}{$\{\vec{x}_{b}^{p},e_{b}^{p}\}$,$\{\vec{x}_{b}^{l},e_{b}^{l}\}$,$P$,$\mathit{firstPhase}$
}} \label{alg:reinit}
\ENDWHILE
\RETURN $\{\vec{x}_{b},e_{b}\}$
\ENDPROCEDURE
\end{algorithmic}
\end{spacing}
\caption{The proposed \DEtl\ algorithm.}
\label{alg:detl}
\end{figure}

The optimization begins with initialization (line~\ref{alg:init} in Fig.~\ref{alg:detl}). Each iteration of 
the $\mathit{while}$ loop (line~\ref{alg:gen}) represents one generation of the evolutionary process. 
In one generation the \texttt{DE/best/1/bin} strategy is performed for each population's vector \{$\vec{x}_{1}$,
$\vec{x}_{2}$, ..., $\vec{x}_{Np}$\} for creating a trial vector $\vec{u}$ (lines~\ref{alg:jde:a} -- \ref{alg:str:b}).
The values of control parameters are set by using the jDE algorithm~\cite{Brest06}.
Each vector is a D-dimensional vector that contains real coded bond $\theta$ and torsional $\beta$ angles:
\begin{equation}
 \vec{x}_i=\{\theta_{i,1},\theta_{i,2}\, ..., \theta_{i,L-2},\beta_{i,1}, \beta_{i,2}, ..., \beta_{i,L-3}\}, \nonumber
\end{equation}
where $D = 2 \cdot L - 5$ is the  dimension of the problem, and $x_{i,j} \in [-\pi,\pi]$. The variable 
$\mathit{firstPhase}$ determines the current optimization phase. The trial vector $\vec{u}$ is evaluated according
to the value of this variable, as shown in line~\ref{alg:if:aux1}. If the trial vector is better than the corresponding 
vector from the population $\vec{x}_i$, then yet another trial vector $\vec{u^{*}}$ is generated using temporal 
locality (lines~\ref{alg:tl:a} -- \ref{alg:tl:b}), and evaluated according to the current optimization phase. 
The second trial vector $\vec{u^{*}}$ is generated by using the promising movement that is added to the best
population vector. In lines~\ref{alg:s1} and~\ref{alg:s2}, the corresponding population vector is replaced by the
better trial vector. The main goal of the first phase is to form good hydrophobic cores, and it is not necessary to
reach very accurate solutions. Therefore, the local search is not used in this phase, as shown in line~\ref{alg:if:local}.
The local search includes a local movement mechanism that allows efficient evaluation of neighborhood vectors which
have moved locally only two consecutive monomers, while all remaining monomers are unchanged. Thus, this mechanism
is only used in the second optimization phase for performing an accurate search.

\begin{figure}[!t]
\scriptsize
\begin{spacing}{1}
\begin{algorithmic}[1]
\algsetup{linenosize=\scriptsize}
\PROCEDURE{\CALL{REINITIALIZATION}{$\{\vec{x}_{b}^{p},e_{b}^{p}\},\{\vec{x}_{b}^{l},e_{b}^{l}\},P,\mathit{firstPhase}$}}
    \IIF{$e_{b}^{p} \le e_{b}^{l}$}{$ \{\vec{x}_{b}^{l},e_{b}^{l}\} = \{\vec{x}_{b}^{p},e_{b}^{p}\}$ } 
    \IF { $\mathit{firstPhase}$ }
        \IF{$\vec{x}_{b}^{l}$ is unchanged for at least $H_c \cdot D$ evaluations within the first phase}
            \STATE {$\mathit{firstPhase} = {\bf false}$}
        \ENDIF
    \ELSE
        \IF {$\vec{x}_{b}^{l}$ is unchanged for at least $P_b\cdot D$ evaluations within the second phase}
            \STATE {// Component reinitialization}
            \STATE $\vec{x}_{i} = $ \CALL{RANDOM}{$\vec{x}_b^l,C$};~~~ $i = 1, 2, ..., \mathit{Np}$ \label{alg:componet}
            \STATE {$ \{\vec{x}_{b}^{p},e_{b}^{p}\} = $ \CALL{BEST}{$P$}}
            \STATE {$\mathit{firstPhase} = {\bf true}$}
        \ENDIF
        \IF{$\vec{x}_{b}^{l}$ is unchanged for $L_b \cdot D$ component reinitializations} \label{alg:cond:componet}
            \STATE {// Random reinitialization}
 			\STATE $\vec{x}_{i} = $ \CALL{RANDOM}{} ;~~~ $i = 1, 2, ..., \mathit{Np}$
 			\STATE {$ \{\vec{x}_{b}^{l},e_{b}^{l}\} = \{\vec{x}_{b}^{p},e_{b}^{p}\} = $ \CALL{BEST}{$P$}}
            \STATE {$\mathit{firstPhase} = {\bf true}$}    
        \ENDIF
 	\ENDIF
\ENDPROCEDURE
\end{algorithmic}
\end{spacing}
\caption{The reinitialization mechanism.}
\label{alg:reinitialization}
\end{figure}

The first generation belongs to the first optimization phase. The reinitialization method is performed
at the end of each generation, and it is responsible for reinitializations and determining the optimization phase. We have defined
three different best vectors~\cite{BoskovicAB18}. The best population vector $\vec{x}_b^p$ is the best vector in the current population,
the local best vector $\vec{x}_b^l$ is the best vector among all similar vectors, and the global best vector 
$\vec{x}_b^g$ is the best vector obtained within the evolutionary process.
How long the local best vector stayed unchanged within the optimization process and the value of control
parameters $H_c, L_b$, and $P_b$, determine the reinitialization and optimization phase.
The algorithm switches to the second optimization phase when $\vec{x}_b^l$ is unchanged for at least 
$H_c\cdot D$ evaluations within the first optimization phase. 
The component reinitialization is performed when $\vec{x}_b^l$ is unchanged for at least $P_b\cdot D$ evaluations within
the second optimization phase, while the random reinitialization is performed when the $\vec{x}_b^l$ is unchanged for
at least $L_b\cdot D$ component reinitializations. In this way, the component reinitialization increases the likelihood
of finding a good similar solution that is different from the already found good solution in only a few components.
The parameter $C$ determines the number of components that are different between $\vec{x}_b^l$ and vectors generated
by component reinitialization (line \ref{alg:componet}). On the other hand, random reinitialization guides the search
process to unexplored search space regions. For a detailed description of all mechanisms of our previous work and its 
influence to the algorithm's efficiency, we refer readers to~\cite{BoskovicAB16} and~\cite{BoskovicAB18}.

\section{Experiments}
\label{sec:experiments}
The \DEtl\ algorithm was implemented by using SPSE (Stochastic Problem Solving Environment), compiled with a GNU C++ 
compiler 5.4.0, and executed using an Intel Core i5 computer with 3.2 GHz CPU and 16 GB RAM under Linux Mint 18.3 Sylvia 
and a grid environment (Slovenian Initiative for National Grid\footnote{Available at \url{http://www.sling.si/sling/}}). 
The grid environment was configured to use AMD Opteron 6272 processor with clock speed of 2.1 GHz, 
cache of 2 MB, and main memory of 128 GB assigned to 64 cores.
The SPSE environment allows for rapid development and testing of stochastic algorithms for different problems in an 
efficient way. The console and web interface are available within this environment. By using the web
interface, we developed a web application that is available at \url{https://spse.feri.um.si}, where the proposed algorithm
can be tested and the optimization process is visualized. In order to evaluate the efficiency of the proposed algorithm, we 
used a set of amino acid sequences, as shown in Table \ref{tab:sequences}. This set includes 18 real peptide sequences 
from the Protein Data Bank database\footnote{Available at \url{https://www.rcsb.org/pdb/home/home.do}}, and 5 Fibonacci 
sequences. The K-D method~\cite{Mount01} is used to transform the real peptide sequences to the AB sequences. In this method,
the amino acids isoleucine, valine, proline, leucine, cysteine, methionine, alanine, and glycine, are transformed to
hydrophobic ones (A), while aspartic acid, glutamic acid, histidine, phenylalanine, lysine, asparagine, glutamine,
arginine, serine, threonine, tryptophan, and tyrosine to hydrophilic ones (B). The selected sequences have different
lengths, which enabled us to analyze the algorithm with the three stopping conditions. The quality of the solution 
($E_t$), or the target scenario, was used for short sequences, while the limited amount of solution evaluation 
($\mathit{NSE_{\mathit{lmt}}}$) and runtime ($\mathit{t_{\mathit{lmt}}}$) were used for long sequences. The used sequences
can also be found in many papers, and, therefore, allow us to compare the proposed algorithm with different algorithms.

Table~\ref{tab:variables} summarizes the expressions that were used in our experiments. Note that all energy values are 
multiplied by $\text{-}1$, which means that all reported energy values are positive, and higher values are better. 
$N_r = 30$ independent runs were performed when the proposed algorithm was compared with the
state-of-the-art algorithms. In all other experiments, $N_r = 100$ independent runs were used. In the
target scenario, experimental results of $\mathit{NSE}$ have near-exponential or near-geometric distribution. Under
such distributions, and with $N_r = N_{succ} = 100$ runs, a reliable rule-of-thumb estimates a 95\%
confidence interval as follows:

\begin{table}[b!]
  \centering
  \small
  \caption{Mean ranks for different settings of the following parameters: $P_b$, $L_b$, $C$, and $H_c$.}
  \label{tab:ranks}
     \begin{tabular}{c|ccrc}
    $\mathit{r_{mean}}$ & $P_b$ & $L_b$ & $C$ & $H_c$ \\
    \hline
    {\bf 3.11} & {\bf 20} & {\bf 20} & {\bf 10} & {\bf 35} \\
    3.11 & 20 & 25 & 10 & 35 \\
    3.22 & 15 & 20 & 10 & 35 \\
    4.11 & 20 & 20 & 10 & 40 \\
    4.44 & 20 & 15 & 10 & 35 \\
    5.33 & 25 & 20 & 10 & 35 \\
    5.78 & 20 & 20 & 10 & 30 \\
    7.11 & 20 & 20 & 15 & 35 \\
    8.67 & 20 & 20 &  5 & 35 \\
    \end{tabular} 
\end{table}

\begin{table*}[t!]
\centering
\small
 \caption{Details of amino acid sequences used in experiments.}
 \label{tab:sequences}
 \resizebox{1.0\textwidth}{!}{
 \begin{tabular}{rrrl}
 label & $L$ & $D$ & sequence \\
 \hline
 1BXP & 13 &  21 & \texttt{ABBBBBBABBBAB} \\ 
 1CB3 & 13 &  21 & \texttt{BABBBAABBAAAB} \\
 1BXL & 16 &  27 & \texttt{ABAABBAAAAABBABB} \\
 1EDP & 17 &  29 & \texttt{ABABBAABBBAABBABA} \\
 2ZNF & 18 &  31 & \texttt{ABABBAABBABAABBABA} \\
 1EDN & 21 &  37 & \texttt{ABABBAABBBAABBABABAAB} \\
 2H3S & 25 &  45 & \texttt{AABBAABBBBBABBBABAABBBBBB} \\
 1ARE & 29 &  53 & \texttt{BBBAABAABBABABBBAABBBBBBBBBBB} \\
 2KGU & 34 &  63 & \texttt{ABAABBAABABBABAABAABABABABABAAABBB} \\
 1TZ4 & 37 &  69 & \texttt{BABBABBAABBAAABBAABBAABABBBABAABBBBBB} \\
 1TZ5 & 37 &  69 & \texttt{AAABAABAABBABABBAABBBBAABBBABAABBABBB} \\
 1AGT & 38 &  71 & \texttt{AAAABABABABABAABAABBAAABBABAABBBABABAB} \\
 1CRN & 46 &  87 & \texttt{BBAAABAAABBBBBAABAAABABAAAABBBAAAAAAAABAAABBAB} \\
 2KAP & 60 & 115 & \texttt{BBAABBABABABABBABABBBBABAABABAABBBBBBABBBAABAAABBABBABBAAAAB} \\
 1HVV & 75 & 145 & \texttt{BAABBABBBBBBAABABBBABBABBABABAAAAABBBABAABBABBBABBAABBABBAABBBBBAABBBBBABBB} \\
 1GK4 & 84 & 163 & \texttt{ABABAABABBBBABBBABBABBBBAABAABBBBBAABABBBABBABBBAABBABBBBBAABABAAABABAABBBBAABABBBBA} \\
 1PCH & 88 & 171 & \texttt{ABBBAAABBBAAABABAABAAABBABBBBBABAAABBBBABABBAABAAAAAABBABBABABABABBABBAABAABBBAABBAAABA} \\
 2EWH & 98 & 191 & \texttt{AABABAAAAAAABBBAAAAAABAABAABBAABABAAABBBAAAABABAAABABBAAABAAABAAABAABBAABAAAAABAAABABBBABBAAABAABA} \\
 \hdashline
 F13  & 13 &  21 & \texttt{ABBABBABABBAB} \\
 F21  & 21 &  37 & \texttt{BABABBABABBABBABABBAB} \\
 F34  & 34 &  63 & \texttt{ABBABBABABBABBABABBABABBABBABABBAB} \\
 F55  & 55 & 105 & \texttt{BABABBABABBABBABABBABABBABBABABBABBABABBABABBABBABABBAB} \\
 F89  & 89 & 173 & \texttt{ABBABBABABBABBABABBABABBABBABABBABBABABBABABBABBABABBABABBABBABABBABBABABBABABBABBABABBAB} \\
 \end{tabular}
 }
\end{table*}

\begin{table*}[t!]
\small
\centering
\caption{Summary of expressions that were used in experiments.}
\label{tab:variables}
\small
\begin{subtable}[t]{0.49\textwidth}
\begin{tabular}{@{}l@{~}p{4cm}}
Expression & Brief description  \\
\hline
 $N_r$ & the number of runs \\
 $E_i$ & the energy value ($E_o$) of $i$-th run \\
 $E_{\mathit{mean}} = \frac{\sum_{i=1}^{N_r}E_i}{N_r}$ & the mean energy value \\
 $E_{\mathit{best}} = \mathit{max}\{E_1, E_2, ..., E_{N_r}\}$ & the best energy value\\
 $E_{\mathit{std}} = \sqrt{\frac{\sum_{i=1}^{N_r}(E_{i}-E_{\mathit{mean}})^2}{N_r-1}}$ & the standard deviation of energy values \\
 $E_\mathit{t}$ & the energy or target value to be reached \\
 $N_{\mathit{succ}}$ & the number of runs where the target value $E_t$ is reached\\
 $S_r = \frac{N_{succ}}{N_r}$ & the success ratio \\
 $\mathit{NSE}_i$ & the number of solution evaluations for $i$-th run\\
 $\mathit{NSE}_{\mathit{mean}} = \frac{\sum_{i=1}^{N_{\mathit{succ}}}\mathit{NSE}_i}{N_{\mathit{succ}}}$ & the mean number of
 solution evaluation for all $N_{\mathit{succ}}$ runs\\
\end{tabular}
\end{subtable}
\hfil
\begin{subtable}[t]{0.49\textwidth}
\begin{tabular}{lp{5.5cm}}
Expression & Brief description  \\
\hline
 $t_i$ & the runtime of $i$-th run \\
 $ t_{\mathit{mean}} = \frac{\sum_{i=1}^{N_r}t_i}{N_r}$ & the mean runtime for $N_r$ runs \\
 $ r_i $ & the setting rank of $i$-th sequence\\
 $ N_s$ & the number of sequences \\
 $ r_{\mathit{mean}} = \frac{\sum_{i=1}^{N_s}r_i}{N_s} $ & the mean rank \\
 $ \mathit{NSE}_{\mathit{lmt}} $ & the number of solution evaluations limit for one run\\
 $ \mathit{t}_{\mathit{lmt}} $ & the runtime limit for one run\\
 $ \mathit{NSE}^{1}, \mathit{NSE}^{2}$ & the number of solution evaluations within the first and second optimization phase\\
 $t^{1}, t^{2}$ & the runtime within the first and second optimization phase \\
\vspace{0.955cm}
\end{tabular}
\end{subtable}
\end{table*}

\begin{footnotesize}
\begin{equation}
\label{confidence}
\begin{split}
\mathit{\mathit{NSE}}_{95} \approx~ & [(1-\frac{1.96}{\sqrt{N_r}})\cdot\mathit{NSE}_{\mathit{mean}} ,\\
 &
(1+\frac{1.96}{\sqrt{N_r}})\cdot\mathit{NSE}_{\mathit{mean}}] \\
 \approx~ & [0.8 \cdot \mathit{NSE}_{\mathit{mean}},1.2 \cdot \mathit{NSE}_{\mathit{mean}}].
\end{split}
\end{equation}
\end{footnotesize}

\begin{table*}[t!]
\scriptsize
\caption{The ranked ($r_i$) control parameter settings ($P_b, L_b, C, H_c$). The population size $Np$ and the 
number of independent runs $N_r$ were set to 100. The stopping conditions were the target energy $E_t$ and
limit of solution evaluations $\mathit{NSE}_{\mathit{lmt}}=2\cdot10^{11}$.}
\label{tab:settings}
\begin{subtable}[t]{0.3\textwidth}
    \begin{tabular}{@{~}c|ccrc|c@{~}}
    $r_1$ & $P_b$ & $L_b$ & $C$ & $H_c$ & $\mathit{NSE_{mean}}$ \\
    \hline
    1 & 15 & 20 & 10 & 35 & 5.371e+06 \\
    2 & 20 & 25 & 10 & 35 & 5.443e+06 \\
    3 & 20 & 15 & 10 & 35 & 5.509e+06 \\
    {\bf 4} & {\bf 20} & {\bf 20} & {\bf 10} & {\bf 35} & {\bf 5.653e+06} \\
    5 & 20 & 20 & 10 & 30 & 5.874e+06 \\
    6 & 25 & 20 & 10 & 35 & 6.076e+06 \\
    7 & 20 & 20 & 10 & 40 & 6.131e+06 \\
    8 & 20 & 20 & 15 & 35 & 6.386e+06 \\
    9 & 20 & 20 &  5 & 35 & 2.151e+07 \\
    \end{tabular}
    \caption{F13, $\mathit{E_t} = 6.9961$}
    \label{tab:f13}
\end{subtable}
\hfill
\begin{subtable}[t]{0.3\textwidth}
    \begin{tabular}{@{~}c|ccrc|c@{~}}
    $r_2$ & $P_b$ & $L_b$ & $C$ & $H_c$ & $\mathit{NSE_{mean}}$ \\
    \hline
    1 & 20 & 20 & 10 & 30 & 5.930e+06 \\
    2 & 20 & 20 & 15 & 35 & 6.079e+06 \\
    {\bf 3} & {\bf 20} & {\bf 20} & {\bf 10} & {\bf 35} & {\bf 6.563e+06} \\
    4 & 20 & 20 & 10 & 40 & 6.790e+06 \\
    5 & 20 & 25 & 10 & 35 & 6.810e+06 \\
    6 & 20 & 15 & 10 & 35 & 7.037e+06 \\
    7 & 25 & 20 & 10 & 35 & 7.055e+06 \\
    8 & 15 & 20 & 10 & 35 & 7.593e+06 \\
    9 & 20 & 20 &  5 & 35 & 2.498e+07 \\
    \end{tabular}
    \caption{1CB3, $\mathit{E_t} = 8.4589$}
    \label{tab:1cb3}
\end{subtable}
\hfill
\begin{subtable}[t]{0.3\textwidth}
    \begin{tabular}{@{~}c|ccrc|c@{~}}
    $r_3$ & $P_b$ & $L_b$ & $C$ & $H_c$ & $\mathit{NSE_{mean}}$ \\
    \hline
    1 & 15 & 20 & 10 & 35 & 1.935e+07 \\
    2 & 20 & 25 & 10 & 35 & 1.937e+07 \\
    {\bf 3} & {\bf 20} & {\bf 20} & {\bf 10} & {\bf 35} & {\bf 2.001e+07} \\
    4 & 20 & 20 & 15 & 35 & 2.014e+07 \\
    5 & 20 & 20 & 10 & 40 & 2.043e+07 \\
    6 & 20 & 15 & 10 & 35 & 2.137e+07 \\
    7 & 25 & 20 & 10 & 35 & 2.288e+07 \\
    8 & 20 & 20 & 10 & 30 & 2.400e+07 \\
    9 & 20 & 20 &  5 & 35 & 9.956e+07 \\
    \end{tabular} 
    \caption{1BXP, $\mathit{E_t} = 5.6104$}
    \label{tab:1bxp}
\end{subtable}

\begin{subtable}[t]{0.3\textwidth}
   \begin{tabular}{@{~}c|ccrc|c@{~}}
    $r_4$ & $P_b$ & $L_b$ & $C$ & $H_c$ & $\mathit{NSE_{mean}}$ \\
    \hline
    1 & 15 & 20 & 10 & 35 & 5.759e+08 \\
    {\bf 2} & {\bf 20} & {\bf 20} & {\bf 10} & {\bf 35} & {\bf 5.835e+08} \\
    3 & 20 & 25 & 10 & 35 & 6.129e+08 \\
    4 & 20 & 20 & 10 & 30 & 6.727e+08 \\
    5 & 20 & 20 & 10 & 40 & 6.840e+08 \\
    6 & 25 & 20 & 10 & 35 & 7.216e+08 \\
    7 & 20 & 15 & 10 & 35 & 8.349e+08 \\
    8 & 20 & 20 & 15 & 35 & 1.317e+09 \\
    9 & 20 & 20 &  5 & 35 & 5.748e+09 \\
    \end{tabular}
    \caption{1BXL, $\mathit{E_t} = 17.3962$}
    \label{tab:1BXL}
\end{subtable}
\hfill
\begin{subtable}[t]{0.3\textwidth}
    \begin{tabular}{@{~}c|ccrc|c@{~}}
    $r_5$ & $P_b$ & $L_b$ & $C$ & $H_c$ & $\mathit{NSE_{mean}}$ \\
    \hline
    1 & 20 & 25 & 10 & 35 & 2.579e+08 \\
    2 & 20 & 20 & 10 & 40 & 2.615e+08 \\
    {\bf 3} & {\bf 20} & {\bf 20} & {\bf 10} & {\bf 35} & {\bf 2.687e+08} \\
    4 & 20 & 15 & 10 & 35 & 2.791e+08 \\
    5 & 15 & 20 & 10 & 35 & 2.806e+08 \\
    6 & 25 & 20 & 10 & 35 & 2.871e+08 \\
    7 & 20 & 20 & 10 & 30 & 2.932e+08 \\
    8 & 20 & 20 & 15 & 35 & 4.820e+08 \\
    9 & 20 & 20 &  5 & 35 & 1.012e+09 \\
    \end{tabular}    
    \caption{1EDP, $\mathit{E_t} = 15.0092$}
    \label{tab:1EDP}
\end{subtable}
\hfill
\begin{subtable}[t]{0.3\textwidth}
    \begin{tabular}{@{~}c|ccrc|c@{~}}
    $r_6$ & $P_b$ & $L_b$ & $C$ & $H_c$ & $\mathit{NSE_{mean}}$ \\
    \hline
    1 & 15 & 20 & 10 & 35 & 4.788e+08 \\
    2 & 20 & 15 & 10 & 35 & 5.141e+08 \\
    3 & 25 & 20 & 10 & 35 & 5.793e+08 \\
    {\bf 4} & {\bf 20} & {\bf 20} & {\bf 10} & {\bf 35} & {\bf 5.972e+08} \\
    5 & 20 & 25 & 10 & 35 & 5.969e+08 \\
    6 & 20 & 20 & 10 & 40 & 6.118e+08 \\
    7 & 20 & 20 & 10 & 30 & 6.230e+08 \\
    8 & 20 & 20 & 15 & 35 & 9.570e+08 \\
    9 & 20 & 20 &  5 & 35 & 2.531e+09 \\
    \end{tabular}
    \caption{2ZNF, $\mathit{E_t} = 18.3402$}
    \label{tab:2ZNF}
\end{subtable}

\begin{subtable}[t]{0.3\textwidth}
   \begin{tabular}{@{~}c|ccrc|c@{~}}
    $r_7$ & $P_b$ & $L_b$ & $C$ & $H_c$ & $\mathit{NSE_{mean}}$ \\
    \hline
    1 & 20 & 15 & 10 & 35 & 1.629e+09 \\
    2 & 20 & 25 & 10 & 35 & 1.649e+09 \\
    {\bf 3} & {\bf 20} & {\bf 20} & {\bf 10} & {\bf 35} & {\bf 1.702e+09} \\
    4 & 25 & 20 & 10 & 35 & 1.833e+09 \\
    5 & 20 & 20 & 10 & 40 & 1.836e+09 \\
    6 & 20 & 20 & 10 & 30 & 1.898e+09 \\
    7 & 15 & 20 & 10 & 35 & 2.089e+09 \\
    8 & 20 & 20 &  5 & 35 & 4.776e+09 \\
    9 & 20 & 20 & 15 & 35 & 8.858e+09 \\
    \end{tabular}
    \caption{F21, $\mathit{E_t} = 16.5544$}
    \label{tab:F21}
\end{subtable}
\hfill
\begin{subtable}[t]{0.3\textwidth}
   \begin{tabular}{@{~}c|ccrc|c@{~}}
    $r_8$ & $P_b$ & $L_b$ & $C$ & $H_c$ & $\mathit{NSE_{mean}}$ \\
    \hline
    1 & 20 & 20 & 10 & 40 & 5.658e+09 \\
    2 & 15 & 20 & 10 & 35 & 6.062e+09 \\
    3 & 25 & 20 & 10 & 35 & 6.458e+09 \\
    4 & 20 & 25 & 10 & 35 & 6.924e+09 \\
    {\bf 5} & {\bf 20} & {\bf 20} & {\bf 10} & {\bf 35} & {\bf 7.366e+09} \\
    6 & 20 & 15 & 10 & 35 & 8.589e+09 \\ 
    7 & 20 & 20 & 10 & 30 & 8.693e+09 \\
    8 & 20 & 20 &  5 & 35 & 1.522e+10 \\
    9 & 20 & 20 & 15 & 35 & 3.296e+10 \\
    \end{tabular}
    \caption{1EDN, $\mathit{E_t} = 21.4703$}
    \label{tab:1EDN}
\end{subtable}
\hfill
\begin{subtable}[t]{0.3\textwidth}
    \begin{tabular}{@{~}c|ccrc|c@{~}}
    $r_9$ & $P_b$ & $L_b$ & $C$ & $H_c$ & $\mathit{NSE_{mean}}$ \\
    \hline
    {\bf 1} & {\bf 20} & {\bf 20} & {\bf 10} & {\bf 35} & {\bf 1.971e+10} \\
    2 & 20 & 20 & 10 & 40 & 2.100e+10 \\
    3 & 15 & 20 & 10 & 35 & 2.145e+10 \\
    4 & 20 & 25 & 10 & 35 & 2.179e+10 \\
    5 & 20 & 15 & 10 & 35 & 2.352e+10 \\
    6 & 25 & 20 & 10 & 35 & 2.396e+10 \\
    7 & 20 & 20 & 10 & 30 & 2.454e+10 \\
    8 & 20 & 20 & 15 & 35 &         - \\
    8 & 20 & 20 &  5 & 35 &         - \\
    \end{tabular}
    \caption{2H3S, $\mathit{E_t} = 21.1519$}
    \label{tab:2H3S}
\end{subtable}
\end{table*}

\subsection{Parameter settings}
Although the two-phase optimization introduces only one new parameter $H_c$, the algorithm works quite differently than 
the previous one. Therefore, in this Section, we will show the influence of the four control parameters $P_b$, $L_b$, $C$, 
and $H_c$ to the algorithm's efficiency while the population size $N_p$ was set to 100 according to the experiment in
\cite{BoskovicAB16}. In our analysis of four parameters, the target scenario ($E_t$) was used on short sequences. 
For each sequence, we used $N_s = 9$ different settings and the obtained results are shown in 
Table~\ref{tab:settings}. Entries that are shown as '-' imply that the algorithm could not reach the
target value $E_t$ in all the runs within the budget of $2\cdot10^{11}$ solution evaluations. The recommended settings and their
results are shown in bold typeface. From the displayed results, we can see that each sequence has its optimal setting, but
it is still possible to select a good setting for all sequences. For this purpose, settings were ranked in 
Table~\ref{tab:settings} and mean rank $r_{\mathit{mean}}$ was calculated for each setting, as shown in Table~\ref{tab:ranks}.
Two settings obtained the best $r_{\mathit{mean}} = 3.11$. We selected the following values
$P_b = 20, L_b=20, C=10$, $Hc=35$, and $N_p=100$ as a good setting, since this setting obtained the best result on the largest
sequence 2H3S. Therefore, this setting is used in all the remaining experiments.
From the displayed results, we can also see that parameter $C$ is the most sensitive parameter. The best rank was 
obtained for the setting with $C=10$, while the worst ranks were obtained for settings with $C=5$ and $C=15$. This can 
also be observed in Table~\ref{tab:2H3S}, where the best result was obtained with $C=10$, while for settings with $C=5$ and $C=15$
the algorithm could not reach the target energy value $E_t$ in all runs. Similar relationships 
of these parameter values can be observed for most sequences in Table~\ref{tab:settings}.

    \begin{table*}[t!]
    \small
    \centering
    \caption{Comparison of the \DElscr, \DElscr$^{*}$, and \DEtl\ algorithms with $N_r = 100$ and 
    $\mathit{NSE}_{\mathit{lmt}} = 10^6$.}
    \label{tab:aux}
    \resizebox{\textwidth}{!}{
    \begin{tabular} {rrrrrrrrrrrr}
    \multirow{2}{*}{Label} &
    \multirow{2}{*}{L} & 
    \multicolumn{3}{c}{\DElscr} & &
    \multicolumn{3}{c}{\DElscr*} & &
    \multicolumn{2}{c}{\DEtl} \\
    \cline{3-5} \cline{7-9} \cline{11-12}
        & & 
        $E_{\mathit{mean}}$ &
        $E_{\mathit{std}}$ &
        {\it p}-value & &
        $E_{\mathit{mean}}$ & 
        $E_{\mathit{std}}$ &
        {\it p}-value & &
        $E_{\mathit{mean}}$ & 
        $E_{\mathit{std}}$\\
    \hline
         F13 & 13 &  4.5648 & 0.9986 & {\bf 6.8001e-27} & &  6.2744 & 0.2525 & {\bf 5.7081e-19} & &  {\bf 6.6551} & 0.2330 \\
        1CB3 & 13 &  6.2329 & 2.0785 & {\bf 4.8569e-17} & &  7.3615 & 0.3759 & {\bf 4.0492e-23} & &  {\bf 8.0308} & 0.3268 \\
        1BXP & 13 &  4.4583 & 0.1995 & {\bf 5.9807e-33} & &  4.9028 & 0.3131 & {\bf 8.9531e-11} & &  {\bf 5.1683} & 0.2028 \\
        1BXL & 16 & 14.5008 & 2.0514 & {\bf 1.6152e-16} & & 15.3518 & 0.4369 & {\bf 8.3531e-18} & & {\bf 15.9518} & 0.4941 \\
        1EDP & 17 & 11.2560 & 2.4148 & {\bf 2.7825e-17} & & 13.1234 & 0.5382 & {\bf 1.9038e-11} & & {\bf 13.6529} & 0.4100 \\
        2ZNF & 18 & 12.5251 & 3.0289 & {\bf 6.1052e-27} & & 15.8738 & 0.5498 & {\bf 8.7550e-14} & & {\bf 16.4476} & 0.7285 \\
         F21 & 21 & 10.0228 & 1.7237 & {\bf 1.9620e-25} & & 13.4101 & 0.9603 & {    0.8517e-00} & & {\bf 13.4152} & 1.3192 \\
        1EDN & 21 & 14.0150 & 3.3468 & {\bf 1.8642e-23} & & 18.0786 & 0.6656 & {\bf 1.1487e-05} & & {\bf 18.4635} & 0.9433 \\
        2H3S & 25 & 12.4696 & 2.5977 & {\bf 1.6222e-24} & & 16.2500 & 1.1672 & {\bf 0.0065e-00} & & {\bf 16.7300} & 1.2836 \\
    \end{tabular}
     }
    \end{table*}

\begin{table*}[t!]
 \centering
 \small
 \caption{Comparison of the following algorithms: \DEtl\ and \DElscr. The shown 
    $\mathit{NSE_{coef}} = \frac{\mathit{NSE_{mean}}(\text{\DElscr})}{\mathit{NSE_{mean}}(\text{\DEtl})}$ and     
    $\mathit{t_{coef}} = \frac{\mathit{t_{mean}}(\text{\DElscr})}{\mathit{t_{mean}}(\text{\DEtl})}$ 
    represent the relationship between corresponding statistics.
}
 \label{tab:twophase}
 \resizebox{\textwidth}{!}{
 \begin{tabular}{rrrrrrr@{~}lrrrrlrrr}
    \multirow{2}{*}{Label} & \multirow{2}{*}{$L$} & \multirow{2}{*}{$D$} & 
    \multicolumn{1}{c}{\multirow{2}{*}{$E_t$}} & \multicolumn{5}{c}{\DElscr~\cite{BoskovicAB18}} & 
    \multicolumn{4}{c}{\DEtl} & \multirow{2}{*}{$\mathit{NSE_{coef}}$} & \multirow{2}{*}{$\mathit{t_{coef}}$} & 
    \multirow{2}{*}{{{\it p}-value}}\\ 
    \cline{5-8} \cline{10-12}
    & & & & $\mathit{NSE_{mean}}$ & $\mathit{NSE_{std}}$ & $\mathit{t_{mean}}$ {\tiny [sec]} &  & & $\mathit{NSE_{mean}}$ & 
$\mathit{NSE_{std}}$ & $\mathit{t_{mean}}$ {\tiny [sec]} \\
 \hline
  F13 & 13 & 21 &  6.9961 & 8.92e+07 & 8.52e+07 &    110.54 &  & & 5.65e+06 & 6.33e+06 &      6.35 &   & 15.8 & 17.4 & {\bf 1.14e-28}\\
 1CB3 & 13 & 21 &  8.4589 & 3.61e+07 & 4.26e+07 &     44.47 &  & & 6.56e+06 & 6.40e+06 &      7.50 &   &  5.5 &  5.9 & {\bf 1.84e-14}\\
 1BXP & 13 & 21 &  5.6104 & 1.56e+09 & 1.68e+09 &  1,965.08 &  & & 2.00e+07 & 1.94e+07 &     22.00 &   & 78.0 & 89.3 & {\bf 3.22e-33}\\
 1BXL & 16 & 27 & 17.3962 & 1.24e+10 & 1.24e+10 & 16,544.45 &* & & 5.84e+08 & 6.06e+08 &    796.22 &   & 21.2 & 20.8 & {\bf 1.66e-30}\\
 1EDP & 17 & 29 & 15.0092 & 4.58e+09 & 4.21e+09 &  7,272.60 &* & & 2.69e+08 & 2.88e+08 &    394.28 &   & 17.0 & 18.4 & {\bf 1.05e-28}\\
 2ZNF & 18 & 31 & 18.3402 & 2.10e+09 & 1.92e+09 &  3,098.81 &* & & 5.97e+08 & 5.73e+08 &    948.52 &   &  3.5 &  3.3 & {\bf 2.36e-13}\\
  F21 & 21 & 37 & 16.5544 &        - &        - &         - &  & & 1.70e+09 & 1.74e+09 &  3,228.19 &   &    - &    - &          - \\
 1EDN & 21 & 37 & 21.4703 &        - &        - &         - &  & & 7.37e+09 & 6.65e+09 & 13,962.33 & * &    - &    - &          - \\
 2H3S & 25 & 45 & 21.1519 &        - &        - &         - &  & & 1.97e+10 & 1.98e+10 & 45,721.34 & * &    - &    - &          - \\
 \end{tabular}
 }
\end{table*}
    
\subsection{Auxiliary fitness function}
An auxiliary fitness function includes knowledge about the quality of the hydrophobic core. This knowledge is crucial
for the efficient optimization process. To demonstrate this advantage of the auxiliary fitness function, we compared
\DElscr, \DElscr$^{*}$, and \DEtl. \DElscr\ (Differential Evolution Extended with Local Search and Component
Reinitialization)~\cite{BoskovicAB18} is our algorithm, that does not have two-phase optimization
while \DElscr$^{*}$ is the same algorithm as \DElscr, but instead of the original fitness function, it uses auxiliary fitness
function throughout the entire optimization process. Only the final population is evaluated with the original fitness function,
and the best vector is returned as a result of optimization. In this experiment,
100 independent runs were performed for each algorithm, and the stopping condition was $\mathit{NSE}_{\mathit{lmt}} = 10^6$.
From the shown results in Table~\ref{tab:aux}, we can observe that the \DElscr$^{*}$ obtained better results in comparison with
the \DElscr\ for all sequences. This demonstrates that the auxiliary fitness function has a good influence on the algorithm's
efficiency. We can also observe that the best results (shown in boldface) are obtained with \DEtl\ which demonstrates that the
two-phase optimization improves the algorithm's efficiency additionally. This conclusion is also supported by
the Mann-Whitney $U$ test. With this test, we verify the equality of the medians between the energy values of the two algorithms.
The shown $p$-values in Table~\ref{tab:aux} represents comparison between of \DElscr\ and \DElscr$^{*}$ with \DEtl. 
From these results, it is evident that the statistically significant difference was identified at the 0.05
level of significance ($p$-value $<$ 0.05) in almost all cases. Only for sequence F21, there was no significant difference between
the results of the \DEtl\ and \DElscr$^{*}$ algorithms.

\subsection{Two-phase optimization}
Two-phase optimization was designed to increase the quality of the hydrophobic cores and, consequently, improve 
the efficiency of the algorithm. In order to demonstrate these advantages, the algorithm with 
two-phase optimization \DEtl\ was compared with the algorithm \DElscr. Within this
comparison, the algorithms were compared by using two scenarios. In the first scenario, the following stopping conditions
were used on the small sequences: $E_t = $ best-known energy value and $\mathit{NSE}_\mathit{lmt} = 10^{11}$.
The results of this scenario are shown in Table~\ref{tab:twophase}, and entries that are shown as '-' imply that the 
algorithm could not reach the target value $E_t$ in all runs. Values marked  with the * are obtained by using
the grid environment, and in these cases $t_{\mathit{mean}} = \frac{\mathit{NSE}_{\mathit{mean}}}{v_{\mathit{mean}}}$.
Here, $v_{\mathit{mean}}$ represents the obtained mean speed of three independent runs on our test computer with 
$t_{\mathit{lmt}} =3600$ seconds. All other results are obtained on our test computer. From these results, it is evident
that \DEtl\ obtained significantly better results in comparison with \DElscr, the distribution of the $\mathit{NSE}$ is
near-exponential or near-geometric, because $\mathit{NSE}_\mathit{mean} \approx \mathit{NSE}_{\mathit{std}}$, and 
\DEtl\ obtained $S_r = 1$ for all sequences with up to 25 monomers, while \DElscr\ only for sequences with up to 18 monomers.
The $\mathit{NSE}_{\mathit{coef}}$ and $t_{\mathit{coef}}$ were calculated for sequences where $S_r = 1$. These coefficients
represent the relationship between the algorithm's results for $\mathit{NSE}_{\mathit{mean}}$ and $t_{\mathit{mean}}$. We can 
see that these statistics of \DEtl\ are decreased from 3.5 to 78 times, and from 3.3 to 89.3 times in comparison with \DElscr.
A statistically significant difference at the 0.05 level of significance for $NSE_{\mathit{mean}}$ can also be observed, 
because 95\% confidence intervals (see Eq.~\ref{confidence}) do not overlap. Additionally, by using the Mann-Whitney $U$ test we
verify the equality of the medians between the $NSE$ of algorithms. The shown $p$-values in Table~\ref{tab:twophase} indicate that
significant difference was identified at the 0.05 level of significance in all the cases. From these results, we can conclude that
the two-phase optimization improves the efficiency of the algorithm for small sequences significantly.

\begin{table*}[t!]
  \centering
  \small
  \caption{The obtained results for \DEtl\ and \DElscr\ within a runtime limit $t_{\mathit{lmt}} = 4$ days for
  $N_r = 100$
independent runs for each sequence.}
  \label{tab:4days}
  \resizebox{\textwidth}{!}{
 \begin{tabular}{rrrrrrrrrrrr}
  \multirow{2}{*}{Label} & \multirow{2}{*}{$L$} & \multicolumn{4}{c}{\DEtl} 
  & & \multicolumn{4}{c}{\DElscr~\cite{BoskovicAB18}} & \multirow{2}{*}{{{\it p}-value}} \\ \cline{3-6} \cline{8-11}
  & & \multicolumn{1}{c}{$\mathit{E_{best}}$} 
    & \multicolumn{1}{c}{$\mathit{E_{mean}}$} 
    & \multicolumn{1}{c}{$\mathit{E_{std}}$} 
    & \multicolumn{1}{c}{$\mathit{S_r}$} 
  & & \multicolumn{1}{c}{$\mathit{E_{best}}$} 
    & \multicolumn{1}{c}{$\mathit{E_{mean}}$} 
    & \multicolumn{1}{c}{$\mathit{E_{std}}$} 
    & \multicolumn{1}{c}{$\mathit{S_r}$} \\
  \hline
  1BXP & 13 & {\bf   5.6104} & {\bf   5.6104} & 0.0000 & {\bf 1.00} & & {\bf   5.6104} & {\bf  5.6104} & 0.0000 & {\bf 
  1.00} & - \\
  1CB3 & 13 & {\bf   8.4589} & {\bf   8.4589} & 0.0000 & {\bf 1.00} & & {\bf   8.4589} & {\bf  8.4589} & 0.0000 & {\bf 
  1.00} & - \\
  1BXL & 16 & {\bf  17.3962} & {\bf  17.3962} & 0.0000 & {\bf 1.00} & & {\bf  17.3962} & {\bf 17.3962} & 0.0000 & {\bf 
  1.00} & - \\
  1EDP & 17 & {\bf  15.0092} & {\bf  15.0092} & 0.0000 & {\bf 1.00} & & {\bf  15.0092} & {\bf 15.0092} & 0.0000 & {\bf 
  1.00} & - \\
  2ZNF & 18 & {\bf  18.3402} & {\bf  18.3402} & 0.0000 & {\bf 1.00} & & {\bf  18.3402} & {\bf 18.3402} & 0.0000 & {\bf 
  1.00} & - \\
  1EDN & 21 & {\bf  21.4703} & {\bf  21.4703} & 0.0000 & {\bf 1.00} & & {\bf  21.4703} &      21.3669  & 0.0431 &      
  0.07 & {\bf 5.76e-36} \\
  2H3S & 25 & {\bf  21.1519} & {\bf  21.1488} & 0.0167 & {\bf 0.96} & & {\bf  21.1519} &      20.9956  & 0.0995 &      
  0.19 & {\bf 2.39e-27} \\
  1ARE & 29 & {\bf  25.2883} & {\bf  24.9863} & 0.1455 & {\bf 0.03} & &       25.2800  &      24.5444  & 0.1718 &     
  0.00 & {\bf 1.59e-31} \\
  2KGU & 34 & {\bf  53.6756} & {\bf  52.9066} & 0.1957 & {\bf 0.01} & &       52.7165  &      51.7233  & 0.3829 &      
  0.00 & {\bf 4.26e-34} \\
  1TZ4 & 37 & {\bf  43.1890} & {\bf  42.7879} & 0.2478 & {\bf 0.03} & &       43.0229  &      41.8734  & 0.4285 &      
  0.00 & {\bf 1.28e-29} \\
  1TZ5 & 37 & {\bf  50.2703} & {\bf  49.7110} & 0.2238 & {\bf 0.02} & &       49.3868  &      48.6399  & 0.3292 &       
  0.00 & {\bf 6.98e-34} \\
  1AGT & 38 & {\bf  66.2973} & {\bf  65.5231} & 0.2948 & {\bf 0.01} & &       65.1990  &      64.1285  & 0.4173 &       
  0.00 & {\bf 6.68e-34} \\
  1CRN & 46 & {\bf  95.3159} & {\bf  93.7138} & 0.5536 & {\bf 0.01} & &       92.9853  &      89.8223  & 0.6514 &      
  0.00 & {\bf 3.36e-34} \\
  2KAP & 60 & {\bf  89.5013} & {\bf  87.6293} & 0.8335 & {\bf 0.01} & &       85.5099  &      83.1503  & 1.0041 &      
  0.00 & {\bf 2.56e-34} \\
  1HVV & 75 & {\bf 101.6018} & {\bf  98.0730} & 1.3038 & {\bf 0.01} & &       95.4475  &      91.4531  & 1.9215 &      
  0.00 & {\bf 2.80e-34} \\
  1GK4 & 84 & {\bf 112.3674} & {\bf 108.2822} & 1.9783 & {\bf 0.01} & &      106.4190  &      99.6704  & 3.0377 &      
  0.00 & {\bf 2.95e-33} \\
  1PCH & 88 & {\bf 166.7194} & {\bf 161.4182} & 2.1279 & {\bf 0.01} & &      156.5250  &     153.1003  & 2.7117 &      
  0.00 & {\bf 3.16e-34} \\
  2EWH & 98 & {\bf 257.0741} & {\bf 250.2833} & 3.1839 & {\bf 0.01} & &      245.5190  &     240.2247  & 2.1421 &      
  0.00 & {\bf 1.95e-33} \\
  \hdashline                                                                                                           
  F13  & 13 & {\bf   6.9961} & {\bf  6.9961} & 0.0000 & {\bf 1.00} & & {\bf   6.9961} & {\bf  6.9961} & 0.0000 & {\bf 
1.00} & - \\
  F21  & 21 & {\bf  16.5544} & {\bf 16.5544} & 0.0000 & {\bf 1.00} & & {\bf  16.5544} & {\bf 16.5304} & 0.0329 &       
 0.65 & {\bf 8.29e-11} \\
  F34  & 34 & {\bf  31.3732} & {\bf 31.2906} & 0.1210 & {\bf 0.10} & &       31.3455  &      30.4913  & 0.3458 &       
 0.00 & {\bf 7.21e-32} \\
  F55  & 55 & {\bf  54.9269} & {\bf 52.7767} & 0.8022 & {\bf 0.01} & &       51.9030  &      49.5009  & 0.8817 &       
 0.00 & {\bf 5.93e-34} \\
  F89  & 89 & {\bf  86.4318} & {\bf 81.3966} & 2.5139 & {\bf 0.01} & &       81.5297  &      76.4804  & 2.0603 &       
 0.00 & {\bf 1.63e-25} \\
 \end{tabular}
 }
 \vspace{0.5cm}
\end{table*}

In the second scenario, the grid environment was used, algorithms were limited with $t_{lmt} = 4$ days, and 
$N_r=100$ runs were performed for each sequence. The obtained results are shown in Table~\ref{tab:4days}.
Both algorithms obtained $S_r=1$ for all sequences up to 18 monomers, while for all other sequences, \DEtl\ obtained better
$S_r$ and $E_{\mathit{mean}}$. From the shown results, we can observe that \DEtl\ obtained a significant improvement in
energy values for longer sequences. For example, $E_{\mathit{best}}$ was improved by 10.1944, 4.9021, and 11.5551 for sequences 1PCH, 
F89, and 2EWH, respectively. Even more, values of $E_{\mathit{mean}}$ that belong to \DEtl\ are better than values of 
$E_{\mathit{best}}$ that belong to \DElscr\ for the following sequences: 2EWH, 1PCH, 1GK4, 1HVV, 2KAP, F55, 1CRN, 1AGT and
2KGU. A statistically significant difference at the 0.05 level of significance for energy values can be also observed
from the shown $p$-values that were calculated by using the Mann-Whitney $U$ test. Note that entries that are shown as '-' imply that
both algorithms obtained the same energy values in all runs and therefore there is no statistically significant difference between the
obtained energy values. From these results, we can conclude that the two-phase optimization improves the 
efficiency of the algorithm significantly for longer sequences too.

\begin{table*}[t!]
    \centering
    \caption{The relationship between the first and the second optimization phase. The shown 
    $\mathit{NSE}_{\mathit{coef}} = \frac{\mathit{NSE}^1}{\mathit{NSE}^1+\mathit{NSE}^2}$ and
    $t_{\mathit coef} = \frac{t^{1}}{t^{1}+t^{2}}$ represent a part of $\mathit{NSE}$ and $t$ used
    by the first phase. Runtime is shown in seconds.}
    \label{tab:relation}
    \begin{tabular}{@{}r|rrrr|l@{~}r@{}}
     Label & $\mathit{NSE^{1}}$ & $\mathit{NSE^{2}}$ & $\mathit{t^{1}}$ & $\mathit{t^{2}}$ 
     & $\mathit{NSE_{\mathit{coef}}}$ & $\mathit{t_{\mathit{coef}}}$ \\
     \hline
     1BXP &    58,128 &    442,080 &   0.146 &    0.980 & 0.116 & 0.130 \\
     1AGT &   857,295 &  9,142,706 &   9.581 &   46.366 & 0.086 & 0.171 \\
     2EWH & 3,704,330 & 96,296,116 & 197.617 & 1269.220 & 0.037 & 0.135 \\
    \end{tabular}
\end{table*}

\begin{figure*}[t!]
\begin{subfigure}[t]{0.32\textwidth}
\centering
\includegraphics[scale=0.32]{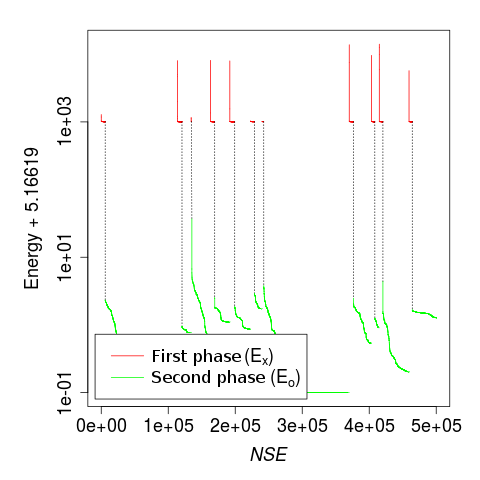}
\caption{1BXP}
\end{subfigure}
\hfil
\begin{subfigure}[t]{0.32\textwidth}
\centering
\includegraphics[scale=0.32]{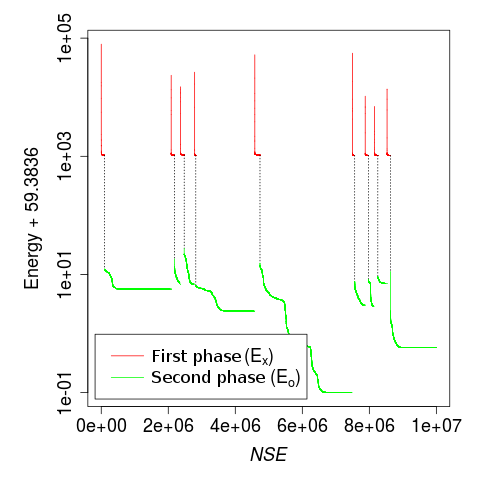}
\caption{1AGT}
\end{subfigure}
\hfil
\begin{subfigure}[t]{0.32\textwidth}
\centering
\includegraphics[scale=0.32]{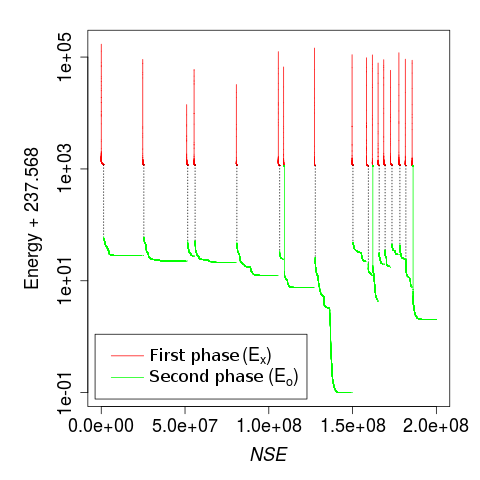}
\caption{2EWH}
\end{subfigure}
\caption{The convergence graphs of the of the best population vector $\vec{x}_b^p$. Some value was added to 
the energy because of the logarithmic scale.}
\label{fig:con}
\end{figure*}

In the continuation of the section, we will analyze both optimization phases and the relationship between them. For this 
purpose, one run was performed for each of the following sequences 1BXP, 1AGT, and 2EWH, with the following stopping 
conditions $\mathit{NSE}_{\mathit{lmt}}=5\cdot10^{5}$, $10^{7}$, and $10^{8}$. The convergence graphs of the best 
population vector $\vec{x}_b^p$ for all three runs are shown in Fig.~\ref{fig:con}. The first phase is shown with the
red line, while the second phase is shown with the green line. The distance between these two lines is determined by
the constant $\lambda$ (see Eq.~\ref{eq:aux}). We can notice that the optimization process is alternated
between two optimization phases, the energy value of the first phase is always higher in comparison with the energy value
of the second phase, and the optimization process is employed mostly in the second optimization phase. This can also be
seen in Table~\ref{tab:relation}. The number of solution evaluations within the first phase $\mathit{NSE}^1$ is significantly
smaller in comparison with the number of solution evaluations within the second phase $\mathit{NSE}^2$. A similar relationship
can be observed for runtime. From the shown coefficients, we can see that the first phase used only 11.6\% of $\mathit{NSE}$
and only 13\% of $t$ for sequence 1BXP. For the longer sequences, these percentages are even smaller. A meticulous reader may
notice that parameters $H_c$ and $P_b$ determine the relationship between optimization phases, and this relationship is
contrary to the parameter values. The reason for this is in the local search, which is used only in the second phase. When
a good solution is reached, the local search is a good mechanism to further improve it. Therefore, the population best vector has
a greater likelihood of improvement and, consequently, the second optimization phase takes more time.

\begin{table*}[t!]
\centering
\small
\caption{Comparison of the \DEtl\ algorithm with state-of-the-art algorithms with $N_r=30$ and
$\mathit{NSE}_{\mathit{lmt}} = M \cdot 10^4$. The displayed $L_2$ represents the percentage of runs where the
second optimization phase is reached. The detailed results are shown in Table~\ref{tab:app:best:alg}.}
\label{tab:best:alg}
\resizebox{\textwidth}{!}{
 \begin{tabular}{rr|rrrrrrr}
 \multirow{2}{*}{label} & \multirow{2}{*}{$M$} & 
 \multicolumn{2}{c}{\DEtl} & \multicolumn{1}{c}{\DElscr~\cite{BoskovicAB18}} & 
 \multicolumn{1}{c}{SGDE~\cite{Rakhshani19}} & \multicolumn{1}{c}{jDE~\cite{Brest06,BoskovicAB18}} & 
 \multicolumn{1}{c}{L-SHADE~\cite{Tanabe14,BoskovicAB18}} & \multicolumn{1}{c}{BE-ABC~\cite{Li15,Li15f}} \\  
 & & $E_{\mathit{mean}}$ & $L_2$ & $E_{\mathit{mean}}$ & $E_{\mathit{mean}}$ & $E_{\mathit{mean}}$ & 
 $E_{\mathit{mean}}$ & $E_{\mathit{mean}}$ \\
  \hline
  1CB3 & 20 &  {\bf 7.3586} & 100.0\% &        4.5108 &       6.0772  &  3.8988 &  2.7916 &  5.9417 \\
  1BXL & 20 & {\bf 15.0934} & 100.0\% &       12.5045 &      14.6894  & 12.4047 & 10.5428 & 11.6942 \\
  1EDP & 20 & {\bf 12.7113} & 100.0\% &        8.1986 &       9.9649  &  7.4667 &  4.5900 &  8.0500 \\
  2H3S & 20 & {\bf 15.1167} & 100.0\% &       11.5310 &      12.6380  & 10.7931 & 10.3830 & 10.4618 \\
  2KGU & 20 & {\bf 38.9910} & 100.0\% &       33.6539 &      38.7383  & 29.5511 & 26.6282 & 22.7195 \\
  1TZ4 & 20 & {\bf 29.8651} & 100.0\% &       21.6863 &      24.1430  & 16.9135 & 16.4693 & 14.9436 \\
  1TZ5 & 20 & {\bf 33.7524} & 100.0\% &       25.9996 &      29.7668  & 20.3655 & 20.6403 & 17.4859 \\
  1AGT & 20 & {\bf 45.7362} &  93.3\% &       39.1897 &      41.4230  & 30.7770 & 29.3564 & 25.6024 \\
  1CRN & 20 & {\bf 69.9021} &  66.7\% &       62.2668 &      64.2589  & 46.9030 & 46.9604 & 42.3083 \\
  1HVV & 20 &      38.2981  &   0.0\% &       35.9335 & {\bf 38.4222} & 20.9541 & 25.4910 & 21.5386 \\
  1GK4 & 20 &      42.0417  &   0.0\% &       42.0261 & {\bf 46.9844} & 22.3218 & 32.9082 & 27.0410 \\
  1PCH & 80 & {\bf 94.6396} &  46.7\% &       87.5748  &           -  & 51.7904 & 59.9509 & 51.6674 \\
  2EWH & 80 &     152.3479  &  16.7\% & {\bf 162.3482} &           -  & 88.8341 & 104.9692 & 94.5785  \\
  \hdashline
  F13 &  4 &  {\bf 4.4955} & 100.0\% &  3.0907 & - &  3.2002 &  2.7742 &  2.8196 \\
  F21 &  4 &  {\bf 9.4729} &  96.7\% &  6.5538 & - &  6.3647 &  5.9441 &  5.2674 \\
  F34 & 12 & {\bf 15.2387} &  96.7\% & 13.3057 & - & 11.5144 & 10.5170 &  8.3239 \\
  F55 & 20 & {\bf 25.6430} &  50.0\% & 22.4019 & - & 16.9941 & 17.1060 & 14.4556 \\
 \end{tabular}
 }
\end{table*}

\subsection{Comparison with state-of-the-art algorithms}
In this section, our algorithm is compared with state-of-the-art algorithms according 
to the number of function evaluations and the best obtained energy values.

\subsubsection{Number of function evaluations}
In the first comparison, the  stopping condition was $\mathit{NSE}_{\mathit{lmt}}$, which was set according to the 
literature \cite{BoskovicAB18,Li15}. The obtained results are shown in Tables~\ref{tab:best:alg}
and~\ref{tab:app:best:alg}. The best-obtained energy values are marked in bold typeface. It can be observed that
\DEtl\ obtained the second best $E_{\mathit{mean}}$ for longer sequences 1HVV, 1GK4, and 2EWH, while, for all the
remaining sequences, it obtained the best $E_{\mathit{mean}}$. Table~\ref{tab:best:alg} additionally shows $L_2$,
that represents the percentage of runs where the second optimization phase has been reached. From these results,
we can see that, for some sequences, \DEtl\ cannot reach the second optimization phase in all the runs because the
value of $\mathit{NSE}_{\mathit{lmt}}$ is relatively small. Some solvers from the literature cannot perform experiments
with a larger value of $\mathit{NSE}_{\mathit{lmt}}$ in a reasonable time because of their runtime complexity.
It is also interesting that, although \DEtl\ did not reach the second optimization phase in any
run for sequences 1HVV and 1GK4, it obtained relatively good results. The SGDE algorithm obtained the best results for
these two sequences. Although SGDE is based on the surrogate model, \DEtl\ outperformed it on all sequences where the
second optimization phase had been reached in most of the runs. For sequence 2EWH, \DEtl\ obtained the second phase in only
5 out of 30 runs, and this can be the reason why it obtained the second best $E_{\mathit{mean}}$ and \DElscr\ the best
$E_{\mathit{mean}}$. When significantly larger number of solution evaluations was allowed with $t_{\mathit{lmt}}=4$ 
days, \DEtl\ outperformed \DElscr\ significantly on all longer sequences, including sequence 2EWH 
(see Table~\ref{tab:4days}).

\begin{table}[t]
\centering
\small
\caption{Comparisons of the best energy values reported in the literature and the best energy values obtained by 
\DEtl.}
\label{tab:best:energies}
\resizebox{0.46\textwidth}{!}{
 \begin{tabular}{rr|rrrr}
  \multirow{2}{*}{label} & \multirow{2}{*}{L} & \multirow{2}{*}{\DEtl} & \DElscr & SGDE & BE-ABC \\
  & & & \multicolumn{1}{c}{\cite{BoskovicAB18}} & \multicolumn{1}{c}{\cite{Rakhshani19}} & 
  \multicolumn{1}{c}{\cite{Li15,Li15f}} \\
  \hline
  1BXP & 13 & {\bf   5.6104} & {\bf   5.6104} &        - &   2.8930 \\
  1CB3 & 13 & {\bf   8.4589} & {\bf   8.4589} &   8.3690 &   8.4580 \\
  1BXL & 16 & {\bf  17.3962} & {\bf  17.3962} &  16.4788 &  15.9261 \\
  1EDP & 17 & {\bf  15.0092} & {\bf  15.0092} &  14.2928 &  13.9276 \\
  2ZNF & 18 & {\bf  18.3402} & {\bf  18.3402} &        - &   5.8150 \\
  1EDN & 21 & {\bf  21.4703} & {\bf  21.4703} &        - &   7.6890 \\
  2H3S & 25 & {\bf  21.1519} & {\bf  21.1519} &  17.3037 &  18.3299 \\
  1ARE & 29 & {\bf  25.2883} &       25.2800  &        - &  10.2580 \\
  2KGU & 34 & {\bf  53.6756} &       52.7165  &  46.0917 &  28.1423 \\ 
  1TZ4 & 37 & {\bf  43.1890} &       43.0229  &  31.5031 &  39.4901 \\
  1TZ5 & 37 & {\bf  50.2703} &       49.3868  &  39.0536 &  45.3233 \\
  1AGT & 38 & {\bf  66.2973} &       65.1990  &  46.2295 &  51.8019 \\
  1CRN & 46 & {\bf  95.3159} &       92.9853  &  78.2451 &  54.7253 \\
  2KAP & 60 & {\bf  89.5013} &       85.5099  &        - &  27.1400 \\
  1HVV & 75 & {\bf 101.6018} &       95.4475  &  52.5588 &  47.4484 \\
  1GK4 & 84 & {\bf 112.3674} &      106.4193  &  57.9654 &  49.4871 \\
  1PCH & 88 & {\bf 166.7194} &      156.5252  &        - &  91.3508 \\
  2EWH & 98 & {\bf 257.0741} &      245.5193  & 201.0500 & 146.8231 \\
  \hdashline                                     
   F13 & 13 & {\bf   6.9961} & {\bf   6.9961} &        - & {\bf  6.9961} \\
   F21 & 21 & {\bf  16.5544} & {\bf  16.5544} &        - &      15.6258  \\
   F34 & 34 & {\bf  31.3732} &       31.3459  &        - &      28.0516  \\
   F55 & 55 & {\bf  54.9269} &       52.0558  &        - &      42.5814  \\
   F89 & 89 & {\bf  86.4318} &       83.5761  &        - &            -  \\
 \end{tabular}
 }
\end{table}

\subsubsection{The best energy values}
\begin{figure*}[h!]
\centering
\begin{subfigure}[t]{0.24\textwidth}
\centering
\includegraphics[trim={0.0cm 0.0cm 3.5cm 0.0cm},clip,scale=0.18]{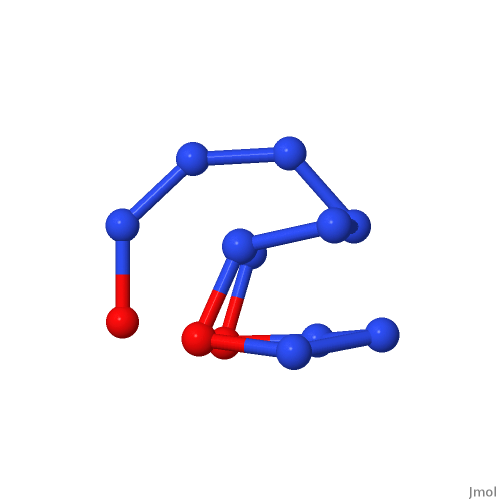}
\caption{1BXP}
\end{subfigure}
\hfil
\begin{subfigure}[t]{0.24\textwidth}
\centering
\includegraphics[trim={0.0cm 0.0cm 2.5cm 0.0cm},clip,scale=0.18]{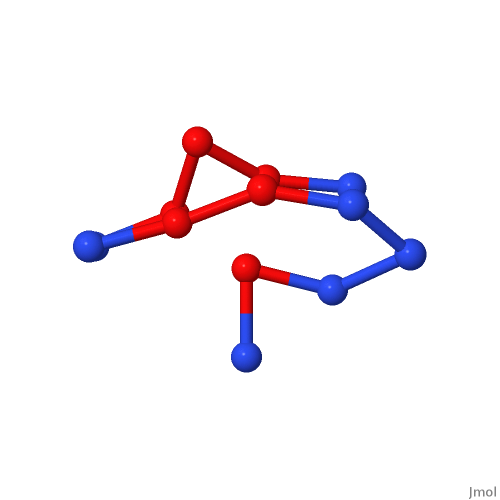}
\caption{1CB3}
\end{subfigure}
\hfil
\begin{subfigure}[t]{0.24\textwidth}
\centering
\includegraphics[trim={0.0cm 0.0cm 3.5cm 0.0cm},clip,scale=0.18]{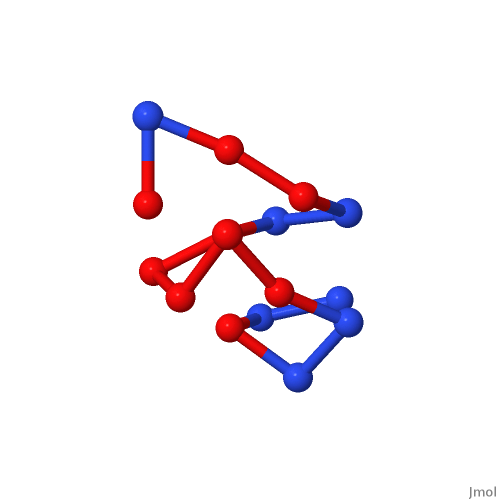}
\caption{1BXL}
\end{subfigure}
\hfil
\begin{subfigure}[t]{0.24\textwidth}
\centering
\includegraphics[trim={0.0cm 0.0cm 3.5cm 0.0cm},clip,scale=0.18]{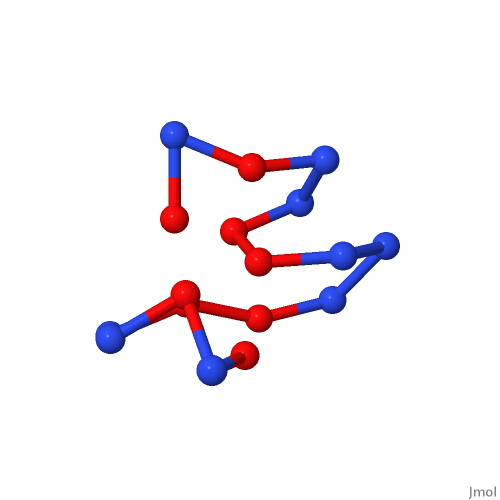}
\caption{1EDP}
\end{subfigure}

\begin{subfigure}[t]{0.24\textwidth}
\centering
\includegraphics[trim={0.0cm 0.0cm 3.5cm 0.0cm},clip,scale=0.18]{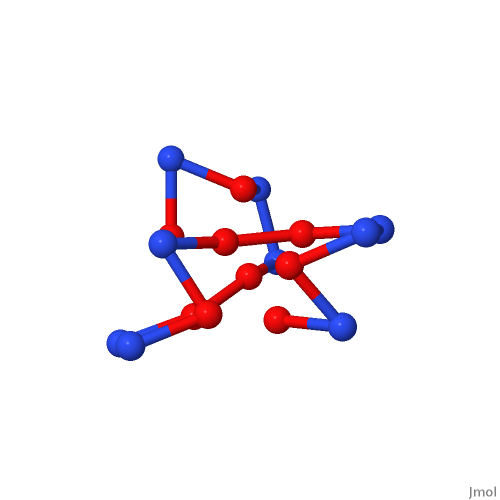}
\caption{2ZNF}
\end{subfigure}
\hfil
\begin{subfigure}[t]{0.24\textwidth}
\centering
\includegraphics[trim={0.0cm 0.0cm 2.5cm 0.0cm},clip,scale=0.18]{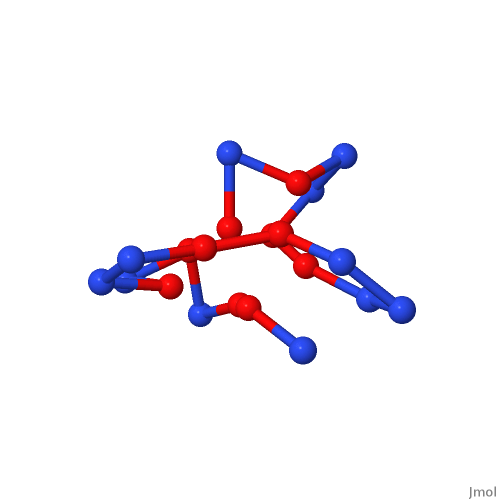}
\caption{1EDN}
\end{subfigure}
\hfil
\begin{subfigure}[t]{0.24\textwidth}
\centering
\includegraphics[trim={0.0cm 0.0cm 1.5cm 0.0cm},clip,scale=0.18]{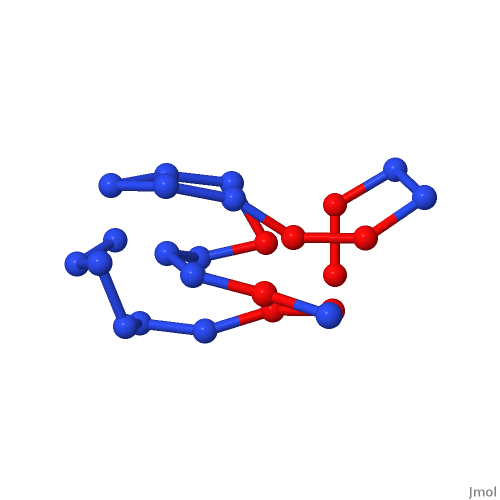}
\caption{2H3S}
\end{subfigure}
\hfil
\begin{subfigure}[t]{0.24\textwidth}
\centering
\includegraphics[trim={0.0cm 0.0cm 1.5cm 0.0cm},clip,scale=0.18]{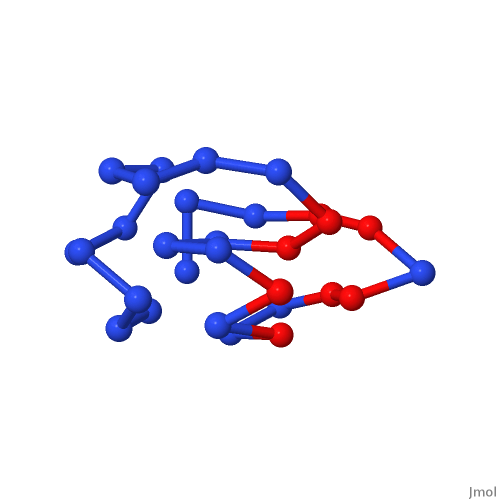}
\caption{1ARE}
\end{subfigure}

\begin{subfigure}[t]{0.24\textwidth}
\centering
\includegraphics[trim={0.0cm 0.0cm 3.5cm 0.0cm},clip,scale=0.18]{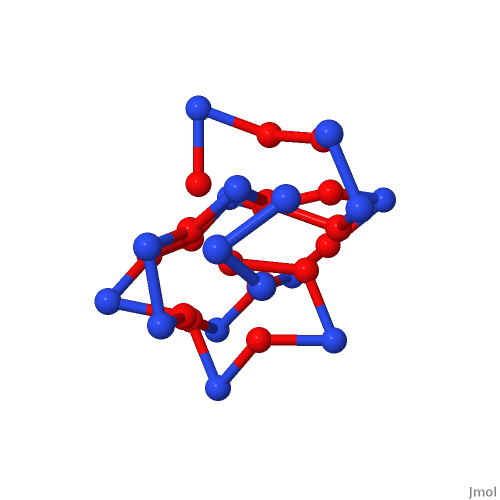}
\caption{2KGU}
\end{subfigure}
\hfil
\begin{subfigure}[t]{0.24\textwidth}
\centering
\includegraphics[trim={0.0cm 0.0cm 2.5cm 0.0cm},clip,scale=0.18]{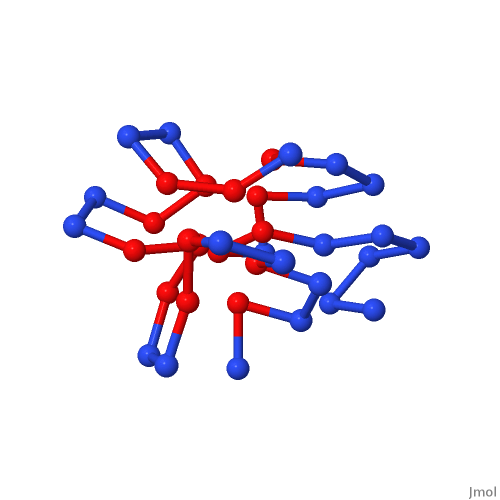}
\caption{1TZ4}
\end{subfigure}
\hfil
\begin{subfigure}[t]{0.24\textwidth}
\centering
\includegraphics[trim={0.0cm 0.0cm 2.5cm 0.0cm},clip,scale=0.18]{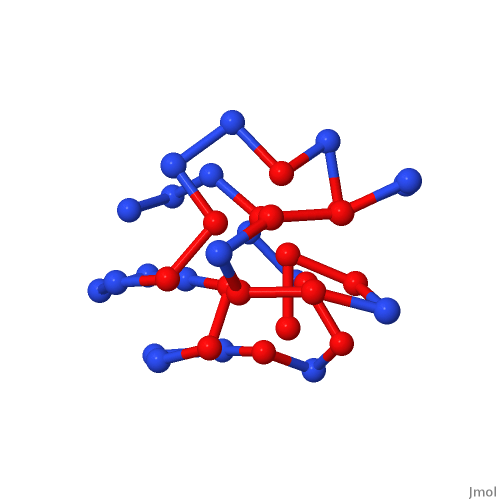}
\caption{1TZ5}
\end{subfigure}
\hfil
\begin{subfigure}[t]{0.24\textwidth}
\centering
\includegraphics[trim={0.0cm 0.0cm 2.5cm 0.0cm},clip,scale=0.18]{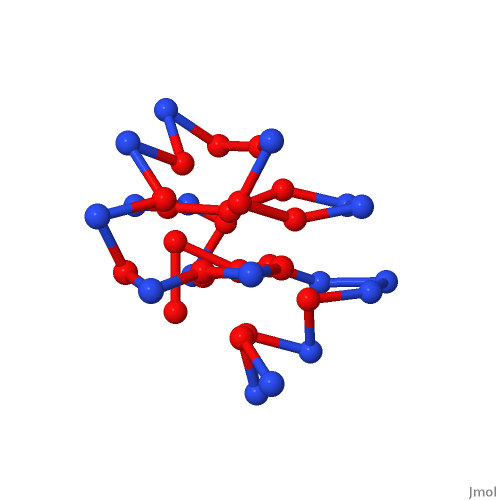}
\caption{1AGT}
\end{subfigure}

\begin{subfigure}[t]{0.24\textwidth}
\centering
\includegraphics[trim={0.0cm 0.0cm 2.5cm 0.0cm},clip,scale=0.18]{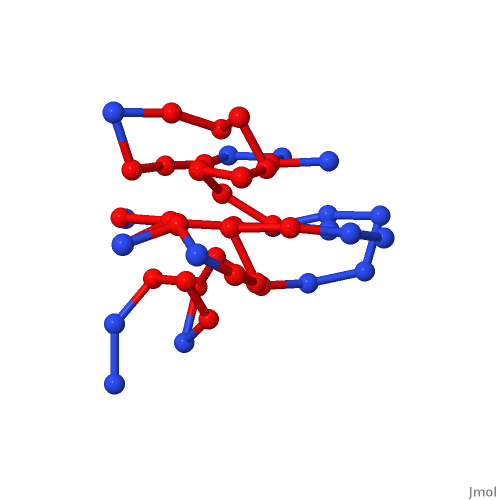}
\caption{1CRN}
\end{subfigure}
\hfil
\begin{subfigure}[t]{0.24\textwidth}
\centering
\includegraphics[trim={0.0cm 0.0cm 2.5cm 0.0cm},clip,scale=0.18]{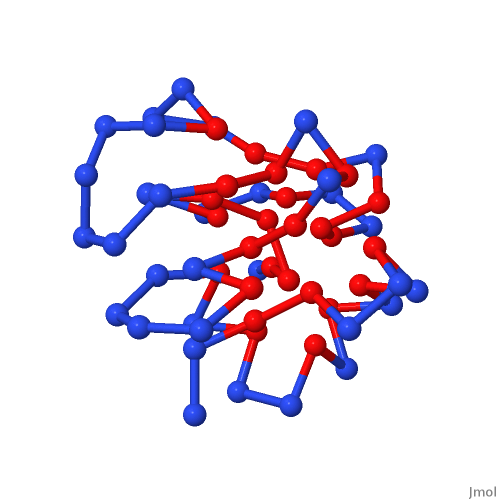}
\caption{2KAP}
\end{subfigure}
\hfil
\begin{subfigure}[t]{0.24\textwidth}
\centering
\includegraphics[trim={0.0cm 0.0cm 2.0cm 0.0cm},clip,scale=0.18]{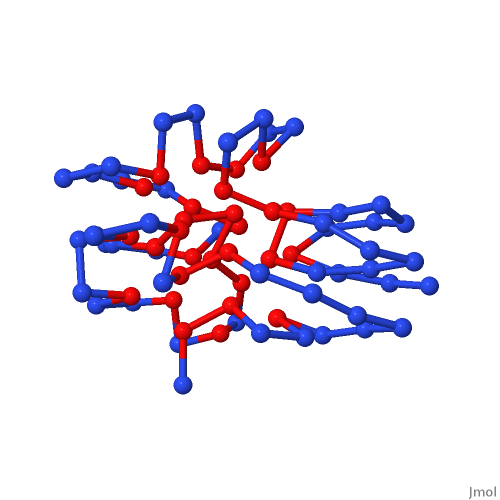}
\caption{1HVV}
\end{subfigure}
\hfil
\begin{subfigure}[t]{0.24\textwidth}
\centering
\includegraphics[trim={0.0cm 0.0cm 2.5cm 0.0cm},clip,scale=0.18]{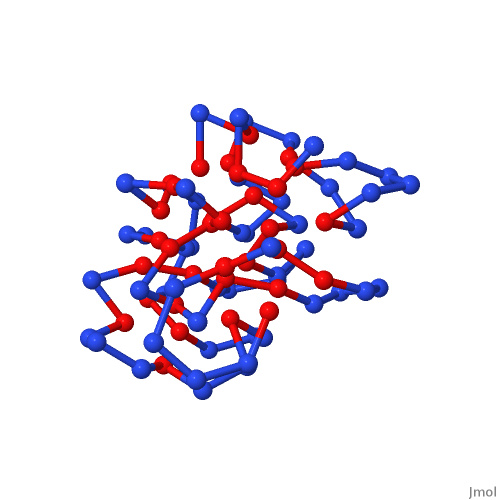}
\caption{1GK4}
\end{subfigure}

\begin{subfigure}[t]{0.24\textwidth}
\centering
\includegraphics[trim={0.0cm 0.0cm 2.5cm 0.0cm},clip,scale=0.18]{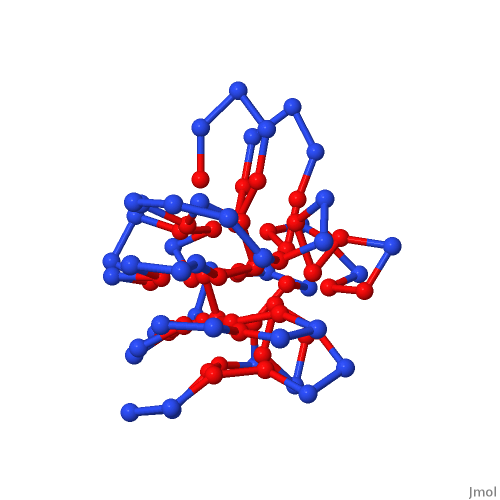}
\caption{1PCH}
\end{subfigure}
\hfil
\begin{subfigure}[t]{0.24\textwidth}
\centering
\includegraphics[trim={0.0cm 0.0cm 2.5cm 0.0cm},clip,scale=0.18]{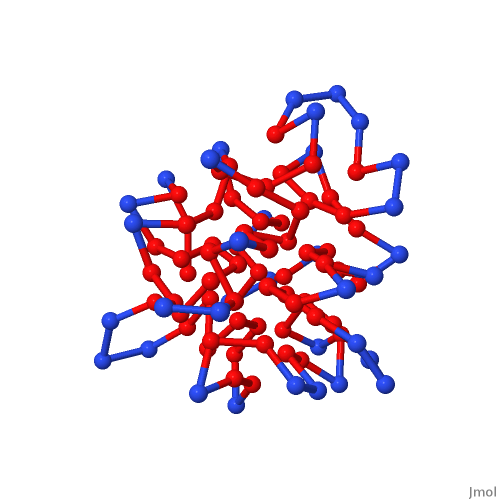}
\caption{2EWH}
\end{subfigure}
\hfil
\begin{subfigure}[t]{0.24\textwidth}
\centering
\includegraphics[trim={0.0cm 0.0cm 2.5cm 0.0cm},clip,scale=0.18]{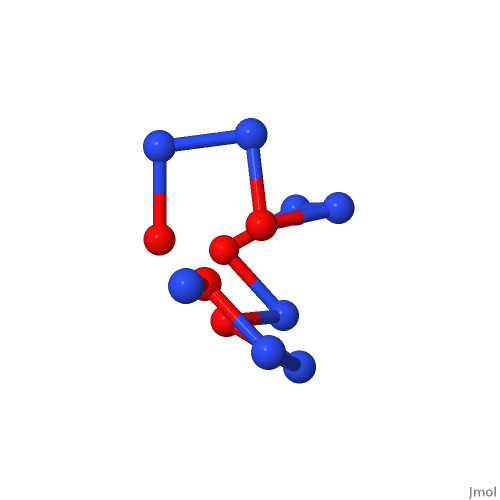}
\caption{F13}
\end{subfigure}
\hfil
\begin{subfigure}[t]{0.24\textwidth}
\centering
\includegraphics[trim={0.0cm 0.0cm 2.5cm 0.0cm},clip,scale=0.18]{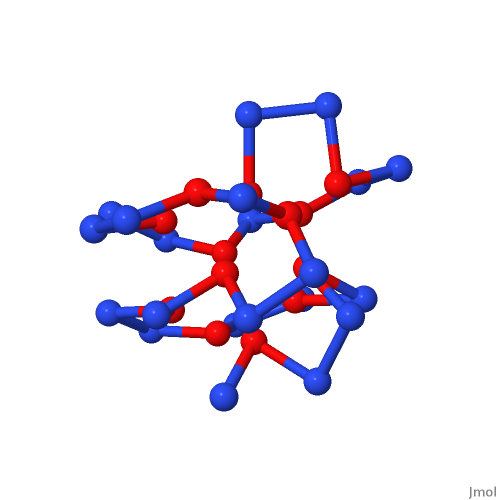}
\caption{F34}
\end{subfigure}

\begin{subfigure}[t]{0.24\textwidth}
\centering
\includegraphics[trim={0.0cm 0.0cm 2.5cm 0.0cm},clip,scale=0.18]{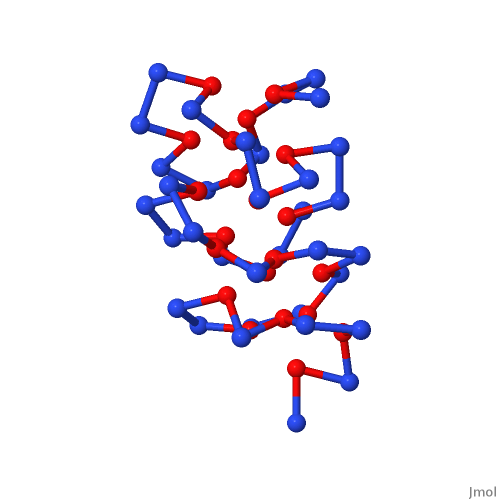}
\caption{F55}
\end{subfigure}
\hfil
\begin{subfigure}[t]{0.24\textwidth}
\centering
\includegraphics[trim={0.0cm 0.0cm 2.5cm 0.0cm},clip,scale=0.18]{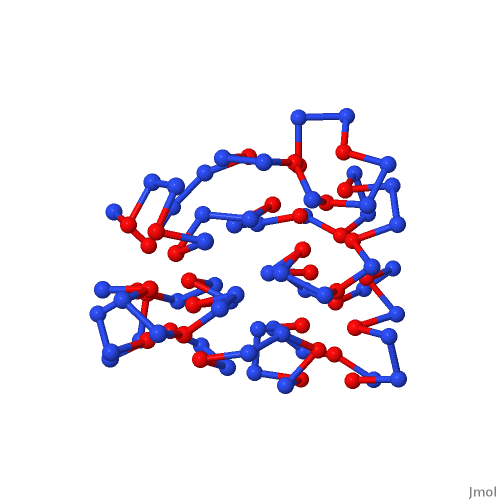}
\caption{F89}
\end{subfigure}
\caption{The best obtained conformation.}
\label{fig:app:solutions}
\end{figure*}

Finally, to demonstrate the superiority of our algorithm in comparison with other algorithms, the best energy values are 
compared for all selected sequences. This comparison is shown in Table~\ref{tab:best:energies}. 
Note that these results were obtained with different experiments. In the case of SGDE, the 30 independent
runs were limited with 200,000 solution evaluations. The running environment was Matlab under Windows 7 operating system 
on an Intel(R) Core i7-6700HQ CPU with 8 GB of RAM. BE-ABC terminated the optimization process when there had been no
evolution for up to 5,000,000 consecutive iterations. In this case, a Matlab environment was used on an Intel
Core 2 Duo CPU with 2.53 GHz and 2 GB of RAM under Windows XP. \DEtl\ and \DElscr\ generated the best solution with the 100 independent
runs that were limited with runtime limit ($t_{lmt}$) = 4 days. In this case, the grid environment was used. 
It was configured to use AMD Opteron 6272 processor with a clock speed of 2.1 GHz, and main memory of 128 GB assigned to 64 cores.
The program was implemented in the C++ programming language and the operating system was Linux.
We can see that \DEtl\ confirms the best energy values for shorter
sequences, and the new best-known solutions were obtained for all sequences with 29 or more monomers.
The solution vectors obtained by \DEtl\ are shown in Tables~\ref{tab:app:solutions1} and
\ref{tab:app:solutions2}, while their graphical representation is shown in Fig.~\ref{fig:app:solutions}. 

\section{Conclusions}
\label{sec:conclusions}
In this paper, we presented two-phase optimization that was incorporated into our Differential Evolution algorithm for
protein folding optimization. In order to improve the efficiency of the algorithm, the optimization process is divided
into two phases. The first phase is responsible for forming solutions with a good hydrophobic core quickly, while the second 
phase is responsible for locating the best solutions. The hydrophobic core represents a set of positions of the hydrophobic
amino acids. Therefore, in the first phase, the auxiliary fitness function is used, that includes expression about the
quality of the hydrophobic core.

In our experiment, we used 23 sequences for analyzing the proposed mechanism and our algorithm for protein structure 
optimization. From the obtained results, we can conclude that the proposed two-phase optimization mechanism improves 
the efficiency of our algorithm. The required runtime for reaching the best-known energy values on small sequences was 
reduced from 3.3 to 89.3 times. In addition, two-phase optimization pushed the frontiers on finding the best-known 
solutions with a success rate of 100\% for sequences from 18 to 25 monomers. The solutions of these sequences could be optimal.
The success rate greater than one is obtained for sequences up to 37 monomers. For these sequences, solutions are close to 
optimal, or could be optimal. For other sequences, solutions are almost surely not optimal, and for these sequences,
the proposed algorithm reached the new best-known solutions.

The proposed algorithm was also compared with state-of-the-art algorithms for protein folding optimization. Although
the used stopping criteria that were taken from the literature did not allow our algorithm to reach the second
optimization phase in all the runs, our algorithm outperformed all competitors on small sequences, and it is comparable on
longer sequences. With the stopping condition of four days, when a significantly larger number of solution evaluations was
allowed, it obtained significantly better energy values for all longer sequences.

In the future work, we will try to implement our algorithm by using full atom and coarse-grained~\cite{Kmiecik16} 
representations of protein structure.

\section*{Acknowledgements}
The authors acknowledge the financial support from the Slovenian Research Agency (research core funding No. P2-0041).

\begin{table*}[t!]
\centering
\caption{Comparison of the \DEtl\ algorithm with state-of-the-art algorithms with $N_r$=30 and
$\mathit{NSE_{\mathit{lmt}}}=M\cdot10^4$. Entries that are shown as '-' imply that no ‘best energy values’ have
been reported in the literature.}
\label{tab:app:best:alg}
\small
\resizebox{0.70\textwidth}{!}{
 \begin{tabular}{rrl|rrrrrr}
 \multirow{2}{*}{label} & \multirow{2}{*}{$M$} & & \multicolumn{1}{c}{\DEtl} & \multicolumn{1}{c}{\DElscr} & 
 \multicolumn{1}{c}{SGDE} & \multicolumn{1}{c}{jDE} & \multicolumn{1}{c}{L-SHADE} & \multicolumn{1}{c}{BE-ABC} \\  
 & & & &  \multicolumn{1}{c}{\cite{BoskovicAB18}} & \multicolumn{1}{c}{\cite{Rakhshani19}} & 
 \multicolumn{1}{c}{\cite{Brest06,BoskovicAB18}} & \multicolumn{1}{c}{\cite{Tanabe14,BoskovicAB18}} & 
 \multicolumn{1}{c}{\cite{Li15,Li15f}} \\
  \hline
  \multirow{3}{*}{1CB3}  & \multirow{3}{*}{20} &
      $E_{\mathit{mean}}$ &  {\bf 7.3586} & 4.5108 &      6.0772  & 3.8988 & 2.7916 & 5.9417  \\
  & & $E_{\mathit{ std}}$ &       0.4137  &   2.13 &      1.7541  & 2.4437 & 2.1068 & 0.7821  \\
  & & $E_{\mathit{best}}$ &       8.0428  & 7.7450 & {\bf 8.3690} & 8.1983 & 8.1151 &      -  \\[5pt]
  \multirow{3}{*}{1BXL} & \multirow{3}{*}{20} &
      $E_{\mathit{mean}}$ & {\bf 15.0934} & 12.5045 & 14.6894 &       12.4047 & 10.5428 & 11.6942 \\
  & & $E_{\mathit{ std}}$ &       0.7750  &    2.17 &  1.8394 &        2.4913 &  2.8712 &  1.1261 \\
  & & $E_{\mathit{best}}$ &      16.5836  & 16.2618 & 16.4788 & {\bf 16.6920} & 14.2015 &       - \\[5pt]
  \multirow{3}{*}{1EDP} & \multirow{3}{*}{20} &
      $E_{\mathit{mean}}$ & {\bf 12.7113} &  8.1986 &       9.9649  &  7.4667 &  4.5900 & 8.0500 \\
  & & $E_{\mathit{ std}}$ &       1.2287  &    2.78 &       2.6239  &  2.9376 &  3.2178 & 0.9330 \\
  & & $E_{\mathit{best}}$ &      14.1823  & 13.1764 & {\bf 14.2928} & 11.9880 & 11.6977 &      - \\[5pt]
  \multirow{3}{*}{2H3S} & \multirow{3}{*}{20} &
      $E_{\mathit{mean}}$ & {\bf 15.1167} & 11.5310 & 12.6380 & 10.7931 & 10.3830 & 10.4618 \\
  & & $E_{\mathit{ std}}$ &       1.7025  &    2.45 &  2.8619 &  2.7864 &  2.6273 &  1.1263 \\
  & & $E_{\mathit{best}}$ & {\bf 18.2800} & 17.1724 & 17.3037 & 16.6920 & 15.6687 & -       \\[5pt]
  \multirow{3}{*}{2KGU} & \multirow{3}{*}{20} &
      $E_{\mathit{mean}}$ & {\bf 38.9910} & 33.6539 & 38.7383 & 29.5511 & 26.6282 & 22.7195 \\
  & & $E_{\mathit{ std}}$ &       3.8462  &    3.99 &  4.6061 &  5.3740 &  2.9071 &  2.0087 \\
  & & $E_{\mathit{best}}$ & {\bf 47.6335} & 41.0221 & 46.0917 & 40.5035 & 35.0707 & -       \\[5pt]
  \multirow{3}{*}{1TZ4} & \multirow{3}{*}{20} &
      $E_{\mathit{mean}}$ & {\bf 29.8651} & 21.6863 & 24.1430 & 16.9135 & 16.4693 & 14.9436 \\
  & & $E_{\mathit{ std}}$ &       3.0009  &    3.62 &  6.1076 &  3.8851 &  2.8963 &  2.2152 \\
  & & $E_{\mathit{best}}$ & {\bf 35.6202} & 34.5265 & 31.5031 & 24.3000 & 20.2216 &       - \\[5pt]
  \multirow{3}{*}{1TZ5} & \multirow{3}{*}{20} &
      $E_{\mathit{mean}}$ & {\bf 33.7524} & 25.9996 & 29.7668 & 20.3655 & 20.6403 & 17.4859 \\
  & & $E_{\mathit{ std}}$ &       3.5719  &    4.12 &  4.5810 &  3.8378  & 3.1163 &  1.3702 \\
  & & $E_{\mathit{best}}$ & {\bf 41.5873} & 37.8896 & 39.0536 & 30.1279 & 34.3115 &       - \\[5pt]
  \multirow{3}{*}{1AGT} & \multirow{3}{*}{20} &
      $E_{\mathit{mean}}$ & {\bf 45.7362} & 39.1897 &      41.4230  & 30.7770 & 29.3564 & 25.6024 \\
  & & $E_{\mathit{ std}}$ &       4.8977  &    5.21 &       6.2854  &  6.3090 &  2.6846 &  2.3415 \\
  & & $E_{\mathit{best}}$ &      53.0663  & 49.9861 & {\bf 54.3623} & 42.9926 & 39.3169 &       - \\[5pt]
  \multirow{3}{*}{1CRN} & \multirow{3}{*}{20} &
      $E_{\mathit{mean}}$ & {\bf 69.9021} & 62.2668 & 64.2589 & 46.9030 & 46.9604 & 42.3083 \\
  & & $E_{\mathit{ std}}$ &       4.8171  &    7.60 &  7.6871 &  7.4243 &  3.7683 &  2.9651 \\
  & & $E_{\mathit{best}}$ & {\bf 78.8733} & 74.7849 & 78.2451 & 63.7138 & 60.2371 &       - \\[5pt]
  \multirow{3}{*}{1HVV} & \multirow{3}{*}{20} &
      $E_{\mathit{mean}}$ & 38.2981 & 35.9335 & {\bf 38.4222} & 20.9541 & 25.4910 & 21.5386 \\
  & & $E_{\mathit{ std}}$ &  4.7919 &    4.92 &       5.9755  &  7.6424 &  1.7090 &  3.5286 \\
  & & $E_{\mathit{best}}$ & 47.6275 & 45.0054 & {\bf 52.5588} & 31.5878 & 28.7787 &       - \\[5pt]
  \multirow{3}{*}{1GK4} & \multirow{3}{*}{20} &
      $E_{\mathit{mean}}$ & 42.0417 & 42.0261 & {\bf 46.9844} & 22.3218 & 32.9082 & 27.0410 \\
  & & $E_{\mathit{ std}}$ &  6.2971 &    4.77 &       3.9699  &  7.4169 &  2.2108 &  3.5287 \\
  & & $E_{\mathit{best}}$ & 55.1498 & 49.9316 & {\bf 57.9654} & 35.6779 & 40.2655 &       - \\[5pt]
  \multirow{3}{*}{1PCH} & \multirow{3}{*}{80} &
      $E_{\mathit{mean}}$ & {\bf 94.6396} &       87.5748  & - & 51.7904 & 59.9509 & 51.6674 \\
  & & $E_{\mathit{ std}}$ &       7.8877  &         11.42  & - & 13.7211 &  3.0935 &  3.4980 \\
  & & $E_{\mathit{best}}$ &     111.0395  & {\bf 121.0579} & - & 83.5786 & 65.9615 &       - \\[5pt]
  \multirow{3}{*}{2EWH} & \multirow{3}{*}{80} &
      $E_{\mathit{mean}}$ & 152.3479 & {\bf 162.3482} & - &  88.8341 & 104.9692 & 94.5785 \\
  & & $E_{\mathit{ std}}$ &  18.7242 &          16.60 & - &  20.2875 &   4.9300 &  5.6967 \\
  & & $E_{\mathit{best}}$ & 192.5963 & {\bf 193.8143} & - & 129.8843 & 118.1532 &       - \\[5pt]
  \hdashline
  \multirow{3}{*}{F13} & \multirow{3}{*}{4} &
      $E_{\mathit{mean}}$ & {\bf 4.4955} & 3.0907 & - & 3.2002 & 2.7742 & 2.8196 \\
  & & $E_{\mathit{ std}}$ &      1.1135  &   0.78 & - & 0.4303 & 0.5559 & 0.3827 \\
  & & $E_{\mathit{best}}$ & {\bf 6.4190} & 4.9533 & - & 4.5359 & 3.8176 & 3.3945 \\[5pt]
  \multirow{3}{*}{F21} & \multirow{3}{*}{4} &
      $E_{\mathit{mean}}$ & {\bf  9.4729} &  6.5538 & - & 6.3647 & 5.9441 & 5.2674 \\
  & & $E_{\mathit{ std}}$ &       1.9427  &    1.53 & - & 0.9800 & 0.7631 & 0.7606 \\
  & & $E_{\mathit{best}}$ & {\bf 13.2876} & 11.1304 & - & 8.7515 & 8.6776 & 6.9065 \\[5pt]
  \multirow{3}{*}{F34} & \multirow{3}{*}{12} &          
      $E_{\mathit{mean}}$ & {\bf 15.2387} & 13.3057 & - & 11.5144 & 10.5170 &  8.3239 \\
  & & $E_{\mathit{ std}}$ &       2.3322  &    2.47 & - &  2.1359 &  1.5477 &  0.9223 \\
  & & $E_{\mathit{best}}$ & {\bf 20.4526} & 19.9550 & - & 15.5885 & 16.6269 & 10.4224 \\[5pt]
  \multirow{3}{*}{F55} & \multirow{3}{*}{20} &          
      $E_{\mathit{mean}}$ & {\bf 25.6430} & 22.4019 & - & 16.9941 & 17.1060 & 14.4556 \\
  & & $E_{\mathit{ std}}$ &        2.5500 &    3.58 & - &  4.0910 &  1.3137 &  1.5594 \\
  & & $E_{\mathit{best}}$ & {\bf 31.2868} & 29.5163 & - & 25.7551 & 21.0993 & 18.8385 \\
 \end{tabular}
 }
\end{table*}

\begin{table*}[t!]
 \centering
 \scriptsize
 \caption{The best solution vectors obtained by the \DEtl\ algorithm.}
 \resizebox{\textwidth}{!}{
 \label{tab:app:solutions1}
    \begin{tabular}{@{}l@{~}p{16.4cm}@{}}
     label & solution vector in degrees $\vec{x}_{b} = \{\theta_{i,1},\theta_{i,2}\, ..., \theta_{i,L-2},\beta_{i,1}, \beta_{i,2}, ..., \beta_{i,L-3}\}$ \\
     \hline
     1BXP & \texttt{ \{     43.2915,    2.8817,  -48.7280,    0.0655,   12.6242,   66.0927,   -6.4080,    8.9633,    8.8002,    2.2354,
                74.0763,     6.6206,   -1.3180,  104.0990, -160.3410,  177.3840,   20.6892,  -26.8003, -127.7890, -166.2700,  -10.2979 \} } \\
     1CB3 & \texttt{ \{    -14.0758,   25.2546,  -38.7358,   -9.5809,   21.0366,   14.7617,   -0.9982,   21.5393,   71.2738,  -27.6012,
                -5.1652,    19.1483,  149.7748, -172.5398, -178.0861, -178.1643,  -91.6772,   -4.8545,   31.1093,  -28.9806,   -3.4154 \} } \\
     1BXL & \texttt{ \{    -22.4292,  -32.2737,  -16.9254,    5.8130,   15.6175,   26.9979,  -38.2372,   52.8361,  -48.2442,  -24.0736,
                49.3335,   -36.1178,   13.9215,   12.5486,   -1.9187,  -55.1452, -147.3023,  127.6298, -168.5915,   62.9624,   27.0891,
               -28.7221,   -27.4283, -152.1219,  177.1523,  -67.7357,    5.2122 \} } \\
     1EDP & \texttt{ \{    -22.6336,    7.2697,   60.7674,   23.9360,  -50.4261,    4.4167,   11.4886,   46.4990,   13.2306,  -12.2668,
                22.7087,     4.0704,   30.6245,  -69.1251,   16.9542,  -26.0209, -124.9106,  155.5754,   61.0880,   -1.5508,  -53.7379,
              -159.4210,   162.5922,  156.4397,  170.4986,   85.1224,   -2.3633,   25.7677,  -67.3571 \} } \\
     2ZNF & \texttt{ \{    -22.5120,    7.7169,  -75.1038,   26.0694,   35.5390,   19.6450,    6.7395,   21.8104,  -57.4641,    1.6924,
                 6.1557,     3.0890,    9.8979,   23.8155,  -48.9192,   -4.3139,  -78.7078,   -2.6658,  114.9430,  148.1870,  162.5640,
                79.1176,    -8.8776,  178.4280,  -42.9368,  -15.8392,   18.6691,  104.1930, -166.4600,  -12.8760, -140.1070 \} } \\
    1EDN &  \texttt{ \{    -23.2048,   31.2207,   46.7641,   48.9338,  -43.6867,  -28.0164,  -17.6723,  -38.3711,  -25.1772,   10.6263,
                 9.0775,    33.5365,   -4.8376,   -6.0992,   25.0580,  -81.1510,   15.5945,   -3.6247,  -36.6783,  -41.0025, -127.4610,
               147.7320,    53.6249,   22.4103,   68.6344,  166.9730, -147.0280,  171.4510,  155.3810, -121.7100,  -29.6786, -131.1440,
               -15.2983,   -24.5428,   54.7787,   83.2637,   29.6805 \} } \\
    2H3S & \texttt{ \{      30.6395,  -51.1362,   34.4028,   -0.4102,  -32.4389,  -10.4102,   -2.0940,   12.4798,   -5.7420,  -60.0843,
                 12.6704,    -8.6855,  -36.5963,  -14.4828,  -17.9173,   13.0795,    0.1480,   17.7335,   -6.0652,    1.4640,  -69.7022,
                  3.0362,    36.2347,   57.1061,  174.6790, -173.2560,  170.6800,  156.7240, -142.5800,  -40.6316,  -22.5668,    1.4454,
               -175.8490,   114.8180,   61.1893,    4.1128,   27.6809,  -84.4735, -144.8670, -176.7310, -161.6050,   97.3255,  158.1730,
               -113.2250,   -54.3451 \} } \\
    1ARE & \texttt{ \{      -11.8099,   -0.1852,  -16.2623,  -42.0892,   19.1083,   -4.8901,   14.4563,   26.9473,    1.1148,  -10.1441,
                29.2761,    -34.4553,   -4.7176,    2.8386,   -3.8010,   33.2357,  -43.3369,   -9.7781,   21.9083,  -19.8608,    4.4000,
                56.1031,     29.8303,    4.6358,  -39.1868,   53.4091,   29.2864,   25.4655,   47.6424,   25.7292,  176.3200, -102.4900,
                -137.3760,  141.2600,   47.0224,  147.7150,   23.0094,  163.2980, -134.0840,  -49.6885,   13.4634,   51.6608,  157.5510,
                -161.3510,  143.8600, -121.9240,  -51.4047, -160.7840,  132.4280,   81.2677,   17.5794, -120.1140,  -67.0551 \} } \\
    2KGU & \texttt{ \{      -20.3903,   -8.0611,   -3.2826,  -67.6941,   45.6930,  -20.2247,  -20.7308,   39.7979,   23.0090,  -80.6396,
                  17.3882,  -11.2956,   49.5839,   22.8948,   48.7441,  -18.5021,   12.0394,   -6.1734,   39.9269,   41.8437,   16.1002,
                  46.2834,  -27.8206,  -67.6604,   51.0005,   -0.5175,  -67.9650,   -6.1230,  -33.1634,   -1.0703,  -40.3086,   36.3314,
                  45.8033,   88.6758,    4.5815, -108.2150, -165.6070,  113.6270, -142.9800,  122.3020,   -9.1593,  -75.3266, -178.6330,
                 -38.7442,  -55.2161,  -38.3169,   59.2034,   25.9876,  -40.6129, -167.2930, -130.5570,  119.4140, -148.7180,  112.9520,
                   4.5520,   29.5852,   -2.8180, -178.3550,  176.0580,   59.9104,   91.2364,  133.9350,   -0.1914 \} } \\
    1TZ4 & \texttt{ \{      -15.7638,   65.4093,  -18.2451,    3.3829,  -16.9888,   -4.9730,   87.4546,   71.4324,  -16.3281,   74.0167,
                  56.2012,    4.7176,  -21.9275,   62.9128,  -23.5336,   35.9449,  -55.6337,   10.3504,  -49.7316,   -6.1732,   32.4112,
                   5.9317,   -4.3330,  -24.1410,   11.9318,   -0.3162,  -82.2922,   -2.1914,   24.2291,  -11.9723,    8.7969,  -15.6866,
                  11.3306,   49.4570,   -6.8914,   50.4939,  172.7930,   68.3833,  156.4490, -118.1980, -154.5940, -174.4790, -108.7440,
                  -4.5956,   28.2398,  161.1620, -170.7260,  -59.2951,   -1.7530,    8.3742, -171.1350,   126.5950,   19.5420,    3.9990,
                   8.6668, -105.5200,   -2.5719,  -48.0064, -151.7840,  165.1080,  -22.5612,  136.9780,    41.1048,  -13.3115,   28.1205,
                 -50.0945, -137.2560,  169.5130,   49.3965 \} } \\
    1TZ5 & \texttt{ \{      -23.1091,  -43.4195,  -14.3960,    0.5110,  -64.3730,   35.2371,    3.6583,    25.6251,   -1.3167,   23.2115,
                 -80.8154,   35.6936,  -46.5693,   36.2667,  -53.9418,   49.8668,    4.2588,   23.5710,    17.1844,   -3.7372,   -8.1432,
                  71.7318,   16.6353,   -5.1681,    4.0692,    2.6351,  -22.0101,   49.0679,  -59.6484,    -4.1104,  -46.0026,   48.8265,
                 -41.1272,   26.0293,   20.4298,   60.3889,  176.5490, -167.3680,  124.7830,  -35.0730,   -19.8600,   -3.4258,  -87.1045,
                -173.8860, -179.2320,  144.4700, -167.6670,  174.9290,   12.0025,  175.2420, -138.9210,  -109.5620,  -46.3340,   54.4235,
                  49.1986,  150.4870, -139.5450,  -96.4876,   -3.2356,   53.7922,  -37.8148,   55.7200,  -168.2500,  -96.8982,  175.9540,
                  74.7548, -161.3030, -130.5270,  128.7780 \} } \\
    1AGT & \texttt{ \{      -24.3263,    5.3764,   10.4396,   -2.1923,   20.4888,  -30.3368,  113.7510,    15.6216,  -56.9395,  -19.0429,
                 -72.2923,  -33.1192,   -6.9467,   10.9927,   62.3836,  -15.9923,   11.3987,  -17.0893,   -12.5705,   21.9368,   -4.2042,
                  -4.3799,    2.7916,  -26.5601,   57.6034,    2.5226,    8.6100,  -18.8326,    1.3634,    24.5203,    6.2251,  -86.2994,
                 -15.2180,  -79.1866,  -75.7196,  -55.4643,   54.0231,  115.4880, -131.4100,  136.0950,  -117.8750,   33.2708,  -24.0852,
                -161.2940,  -38.9457, -148.5660,  -25.2700,   51.9177,   80.9154, -162.7170,  -34.4541,   -11.0239,  -99.0138,  171.8190,
                 148.9170, -158.1880, -122.4920,  -37.4450,   48.1155, -164.7280,   78.8424,   -2.8201,   -35.0597,    7.5444,  109.2200,
                 157.7730,  -38.3773,  169.1840,   34.7763,  147.2250,   44.1010 \} } \\
    1CRN & \texttt{ \{       50.0786,   -4.0572,  -61.2308,   44.6842,   76.8231,   57.6188,  -32.4502,     2.4721,   12.0211,   74.3335,
                  -0.8714,   12.0442,   -0.6223,   -2.6659,   -5.0746,   -1.1086,   -5.0429,   27.4888,     6.1272,   19.9300,  -55.2943,
                 -28.5681,  -23.0060,  -64.9208,    0.5466,   -5.4714,   -6.4818,   -8.4072,   -4.9637,   -32.0666,  -44.4978,  -12.3099,
                  24.4822,  -61.0707,   23.1750,  -15.7684,    1.2729,  -72.7992,   13.3133,    5.0746,    26.3956,   -3.4537,   34.6629,
                   2.6228,   54.3496,  -43.1639, -127.9010,  -12.0315,   52.1853,  -24.4792,  -34.9035,     0.9760,   -0.5159, -151.7190,
                -171.1650,  133.8860,  164.7700, -156.5570,   90.6416,   30.7998,  147.5420,   85.8871,   -25.4746,   55.2350,  -48.5308,
                 -68.2493,  169.7940,   16.1425,  -10.2563,  -52.5471, -154.8070,  164.0250, -178.9780,   120.0000,   44.7010,  -63.6278,
                 159.3880, -109.0940,  146.7800,  157.5190,  -15.0846,  -59.3834,  -49.8460,   -63.0123,    18.6074,  105.8140,   -3.9525 \} } \\
    2KAP & \texttt{ \{       24.3718,   27.4667,  -44.0114,   42.1294,  -59.6325,  -44.5874,    -5.5296,   -31.0558,    3.8587,  -74.0998,
                 -37.9655,   68.1791,  -11.7538,   72.6772,   -5.4661,   63.9122,    2.6613,    44.6125,   -21.2170,   -1.4292,   39.5007,
                 -17.5291,   21.1484,   27.0784,   53.4574,   18.9653,  -52.7529,   59.6967,    16.2715,     8.8282,   46.2614,  -12.4821,
                  86.2839,   71.9082,   -1.7970,   -2.0548,  -51.5120,   37.8161,   -7.8908,   -34.5328,   -15.1049,   14.2872,  -90.8831,
                  23.9633,  -29.2878,   13.1950,  -42.4110,    7.1498,  -14.1595,   20.3454,   -18.4859,   -18.9042,   -2.1230,   -4.0425,
                 -20.7228,  -70.4339,  -36.7226,   17.3609,   19.0532,   43.6405,   14.0840,  -127.4210,   -22.1486, -144.8530,  -50.4490,
                -169.6760,   11.7031,  131.7470,  179.9640, -146.4880,   -8.9112, -177.3020,    22.4927,   160.3420,  144.9730,   65.3899,
                  23.8940,  -20.6934, -151.8260,  -26.5497,   43.2933,   38.2212,  -71.2860,   175.9890,  -148.9180,  140.4420,  166.3260,
                -170.7830, -122.6240,   21.6178,  -36.6286,   39.6738,  -22.0121, -147.5460,  -129.4730,   -13.6816,   36.4449,    9.4886,
                 -13.4387,   40.1758,  166.4200,  -76.3177,  -45.6436, -146.6620,  141.4190,  -114.9160,   168.8570,   76.1944,  179.5190,
                  96.0169,    6.7884,  -24.1243,    9.0689, -108.3530, -102.5490 \} }\\
     1HVV & \texttt{ \{         28.2685,  -40.6918,   -7.7275,  -32.3201,  -20.7837,   10.2414,     6.1012,  -15.4520,  -26.0580,  
                 -21.1452,  -33.2152,   25.2166,   22.1740,   73.0179,  -12.3503,   13.0662,    2.7632,     1.3007,    4.7266,  -85.3002,   
                  -5.4729,   14.8792,    2.8751,    4.0576,  -84.8795,   15.0615,   30.9497,   82.3684,   -38.9026,   63.6890,    7.7874,   
                  41.7187,   -4.1016,    0.5837,   -7.2732,   -9.6488,   44.9434,   12.2383,  -21.3734,   145.9480,  -11.3667,  -19.0659,
                 -14.8583,   15.8618,   12.0296,  -10.1809,   85.8912,   20.7189,  -84.4999,   -4.8766,    49.9831,    4.9592,   47.0089,
                  87.6569,   36.3121,  -95.9835,  -22.6773,  -13.3727,  -31.4584,  -13.4276,   10.6063,     3.6482,  -15.5338,   69.6840, 
                  -2.2884,   10.2172,  -32.7510,   -5.2629,    7.0523,   35.9366,  -21.7282,  -11.0987,    -3.2714,   43.7507,  -33.8599,
                -127.1520,  -20.7062,   38.4249,   50.7762,  152.1700,  156.1180,  179.5900, -132.1530,  -171.4480,  106.2230,  -12.4394, 
                 120.1950,  165.4130,  -93.5222,  -56.1465, -163.2590,   16.8057,  -29.1228, -137.3300,   -49.7135,   35.8770,  -30.4969,   
                  31.4916,  -70.1748,   37.9770,  136.5180,   14.9668,  159.6950, -138.9450, -148.1880,  -101.2000,   -7.5629,   41.1898,
                 -42.7400,   67.5312,  161.6920,  -49.0135,  147.9620,   59.5970,  163.6870,  117.7390,    24.2642,  -30.7786,   15.5388, 
                 -53.0329,  -22.0126,  -49.6826,  -17.9750,   60.3983,  155.4430,   42.1223,  142.8740,   -33.7056,  -11.3703,   23.0922,
                 16.8769,   -30.8626, -138.3300,  175.4940, -154.8640,  -16.5083,   12.0973,   42.3729,   -53.0223, -120.5350, -153.5300,  
                 152.0010,   17.4168,   -3.2292,   46.0565 \} } \\
      \end{tabular}
     }
\end{table*}

\begin{table*}[t!]
 \centering
 \scriptsize
 \caption{The best solution vectors obtained by the \DEtl\ algorithm.}
 \label{tab:app:solutions2}
  \resizebox{\textwidth}{!}{
     \begin{tabular}{@{}l@{~}p{16.4cm}@{}}
     label & solution vector in degrees $\vec{x}_b = \{\theta_{i,1},\theta_{i,2}\, ..., \theta_{i,L-2},\beta_{i,1}, \beta_{i,2}, ..., \beta_{i,L-3}\}$ \\
     \hline  
 1GK4 & \texttt{ \{        -23.5368,  -52.9765,  -91.0886,  -22.5084,   48.5852,   46.2424,     6.7747,  -18.6429,  -29.4579,   12.3254,
                  32.9086,   -9.8384,    0.3103,  -57.7047,  -35.1338,   79.5519,  -21.6135,    70.1648,  -52.2448,  -28.3012, -137.8700,
                   3.0760,  -24.7032,   30.5878,  -33.0627,    8.8434, -125.1280,  -24.2820,    41.7009,   49.2211,   11.0426,  -18.3751,
                 -31.3209,   -9.8915,   15.8866,  -53.1167,   23.5343,   -6.1429,  -33.0249,     1.3760,  -29.4937,   26.8502,  -69.2160,
                 -31.5893,    9.5558,   15.2941,  -35.7348,  -44.0274,   12.8923,   65.1494,   -17.7063,  -10.1409,   -3.2489,  -13.8772,
                  81.8159,  -30.3965,   -0.8736,   73.0233,  -37.7217,   -4.2948,  -19.1917,   -41.4329,   10.1900,   31.4222,   55.1983,
                 -25.3248,  -33.2248,   57.2687,  -10.9687,  -23.7223,   20.4717,    8.1177,     6.1580,  -14.6142,  -33.8161,  -17.7138,
                  23.7175,   22.1855, -110.1050,  -42.2455,   15.3911,   67.9538,  108.5460,    -8.2056,   40.3298,   -4.8811, -117.2820,
                 -11.8158,  -31.9969,   69.1024,  140.0150, -165.8220,  -50.0041,  -87.3274,  -161.6650,  162.0850,  -67.1328,  171.2760,
                 151.3160,  -72.9197,  -14.4334,  -16.9347,   94.3613,  126.1540,   28.3020,    -2.0875,  123.2920,  -40.9864,  -50.1576,
                 -40.0075, -146.2250, -169.6290,  100.8440,   123.098,  159.8380,  164.4800,    -9.9628,  129.1940,   77.8586,  -23.1678,
                 -70.0338,   32.3327,  -26.1254, -155.4590,  -42.4838,  -97.7798,  160.0390,   132.2410, -149.6070,  -73.8934,    0.8318,
                 119.7600, -132.7580,  -73.1846,  -21.7379,   20.8450,  123.8960,   86.9948,   123.1030,  140.3210,   22.3489,   76.3421,
                 -33.8403, -101.6840,  150.6400,  155.2420,  -49.0566,   58.1313,  -24.0440,   -15.2747,  -41.5244,  -56.1580,   45.9663,
                 108.4570,  144.3050,  172.4450,   96.9614, -150.6490,  160.6200,    1.8834,   -15.2365,  -20.9220,   -6.8423 \} } \\
  1PCH & \texttt{ \{         44.5556,  -54.2356,  -99.8784,   63.3900,   79.3698,   80.0025,    36.2528,  -63.5092,   69.2080, -102.3630,
                 -46.4701,   12.8899,  -33.9664,  -45.0167,   21.7445,   -4.8924,  58.7080,    -9.3451,   50.1629,  -71.1396,   39.1343,
                 -93.8616,   23.7380,   -9.5606,  -48.9556,  -13.8713,   -4.7244,   -5.5232,   -24.5021,   56.6785,  -24.8860,  -72.7329,
                  14.6568,  -32.6986,   27.8478,   -6.2063,   13.5359,  -23.1337,   -1.0471,   -10.5722,  -26.8637,   62.2722,  -18.5086,
                   4.7446,   23.5224,  -58.3009,    9.3932,    6.2841,  -74.2319,   -4.9259,    11.6023,  -21.3455,   10.8119,   -6.6138,
                 -14.8199,   -1.9473,   -7.7355, -153.1700,   28.1558,   34.6185,   32.6096,    10.3379,   13.9097,  -18.7154,   73.1461,
                  -7.6761,    2.2450,  -19.7779,  -10.4513,   49.0782,  -88.2309,   -4.0134,    19.5760,   45.2757,   -0.0988,    9.6584,
                  37.5556,    4.6149,  -28.0496,   31.4917,  -53.0738,  -30.8891,   75.8167,   -44.1626,   77.1048,   21.4111,   -2.4662,
                -154.5360,  -40.5056,   -9.1083,   14.6453,    8.4616,  173.0740,    4.2692,  -119.6730,  -46.9798,   58.6207,   -8.9283,
                 132.1090,   50.0801,   -3.0832, -168.3060,  149.2040,  174.0960,   48.1556,    38.6760, -149.7110,  107.5950,  157.9690,
                -174.9360, -138.3730,  -85.4430,  -23.4847,   61.7108,  -33.9020,  -30.0635,   134.6030,  116.7520,  104.9670, -155.3420,
                 -94.9446,  -55.4257, -159.9640,  -46.2382, -151.2140,   -1.4016,   20.0691,   -52.3698, -151.1780,   41.8111,   19.4301,
                 130.6930,  -27.8527,   73.8315,   23.8190,   44.5917,  134.1470, -136.3670,   117.8080, -160.5790,  -68.0745,  -17.0275,
                 -47.7369, -155.1620,  -41.3912, -108.3610,    5.2759, -100.9010,    2.1443,    54.7888,  -52.8865,   27.3697,  117.8370,
                 178.2460,  -28.0998,  135.4250,  100.7990,  175.9960,  138.6450,   36.4198,   -13.5205,  -16.0751,   27.6616,  -32.6852,
                -148.3610, -156.2740, -176.0750,  -26.7138,   -9.1947 \} } \\
  2EWH & \texttt{ \{        116.1390,  -49.4772,   10.3899,  -58.2751,  -23.3784,   24.6000,   -36.7045, -149.0510, -132.7830,  -21.0727,
                  23.5253, -104.1120,   13.1179,   30.4175,   50.7226,  -92.4012,   42.7474,   -20.1564,   52.6708,  -88.4301,   56.8142,
                 -27.6274,   18.5940,   76.6203,   -3.5028,  -52.0687,  -21.2880,  -49.3951,   -31.6283,  -32.7019,   87.3416,   35.5355,
                 153.5040,   55.1270,  -38.5958,  -44.1908,  -60.1665,  124.9320,  -32.6850,   -36.4796,  -24.9186,  -31.4726,   16.5334,
                 -52.5521,  -14.6882,   -8.9236,  -31.3533,    1.2030,   15.0014, -147.0140,    17.5299,   29.0412,   22.3586,   52.5274,
                 130.4610,   46.9053,  -17.6291,   -2.6997,  -40.4363,  103.5650,  -64.7322,    72.7882,   26.9161,   -1.2370,  -70.9261,
                 -49.9822,   20.7601,   -4.3949,   33.2222,  116.6310,   19.3606,  -13.3266,   -32.9903,  -10.2704,   70.8096,  -26.2030,
                 -31.8590,  -44.9023,  -19.0768,   26.1096,   86.7503,   28.3688,   62.0058,    10.8096,  -55.6498,  -94.7941,   11.2196,
                -100.2350,    9.5840,  -24.8805,  -46.3249,   34.0227,  -71.2436,  -51.6366,   -31.1617,   35.2951,   63.5008,  178.0460,
                  -8.7774,   53.6840,  -50.0632,   50.8550,   53.7114,  -30.4684, -124.7150,  -164.8180,  -33.5956,  -23.1001,   24.0731,
                  41.1648,   16.1909,   35.4786,  -66.5841, -138.5830,    3.7838,   54.9703,   111.8690,  134.3130,   -2.6298,    1.0777,
                  20.5968,  -62.7794, -151.8680,  -69.6083,  -22.6480,  -12.1114, -120.8450,    34.6737,   43.1051,   47.5810,   18.0300,
                  61.4490,   16.3383, -152.8790, -131.6510,  113.7650,   55.8909,   15.1288,  -125.2830,  -15.5483, -130.0730, -131.6170,
                 -63.4486,  -99.5910,   32.9298, -107.2520,  164.8250,   80.9774,   41.5560,   -48.8652,  -36.5543,   69.0207,  115.5430,
                -136.8630,   -1.0986,   64.7935,  157.8040,  125.2570,  164.9310,  -13.8925,    44.7431,   98.2381,    6.9066,  -72.9673,
                  -0.7786,   55.2588,  -55.9858, -122.9650,  -31.8611,   52.8912,  155.6490,   168.2810,  -76.7438,  -24.3165,   12.0668,
                  -4.6317, -147.7810,  -35.5362,  -29.7130,   42.1289,   11.3713,   36.2228,    29.0703, -158.5470, -133.9110, -141.0250,
                 -36.0370,   14.3460,   38.1235,   -6.7165, -171.0050 \} } \\
  F13 & \texttt{ \{           7.6652,  -83.4480,   13.0886,    0.5513,   29.1616,  -47.9080,     2.7533,  -31.0327,  -31.3119,  -46.3918,
                   0.2762,    9.0488,  -29.5745, -116.1991,  160.5075,    0.8902,  129.3809,    24.5074,  113.3802, -161.6724,   98.7127 \} } \\
  F21 & \texttt{ \{          -5.7082,  -70.6345,   12.6013,  -78.4561,    5.1401,    2.4915,    57.5974,  -25.4160,   27.2287,  -35.8677,
                  -5.3343,  -13.9895,    3.0216,   19.9055,   74.4006,  -31.0708,    4.7647,   -19.1022,  -32.9492, -155.5060,   16.0013,
                 169.1010, -162.8930,   94.9124, -155.5030,  140.8910, -153.3320,  -40.6752,  -137.5630,  -48.1957,   35.2245,  -66.7533,
                  37.5734, -137.9090,  144.5210,   52.7295,  156.8710 \} }\\
  F34 & \texttt{ \{           6.5328,  -83.0367,   15.1104,   16.9355,   28.8433,    5.2647,    52.5152,  -13.0130,  -25.2523,   -8.0214,
                   7.0780,   11.7256,   22.0270,   -9.2043,  -19.2205,  -67.3482,   35.1195,   -61.2379,   31.1857,   11.0780,    4.1848,
                 -27.4726,   -1.3645,   17.3948,   21.7434,   -3.2610,    2.2779,  -27.9407,   -48.4669,   65.0824,  -31.0953,   60.5105,
                   7.7212,  -33.0859, -119.1020,  154.2810, -130.4210,  124.2230, -143.5030,   138.7690,   43.7392,  147.0940,   61.7037,
                 -26.8124,   57.2326,  -54.5721,   42.5337, -159.4070, -126.2040,  164.1620,   -82.2574, -146.6720,  -55.1973,   26.5960,
                 -75.9919,    8.7552,   97.1129,   29.5944,  148.8120,   38.8499, -155.1420,  -157.7690,  138.1670 \} } \\
  F55 & \texttt{ \{         -14.6178,  -81.6545,   19.6900,   -5.6211,  -11.7680,   22.8494,   -69.8362,   14.4445,  -43.1447,   -3.8896,
                   1.1940,   15.5990,    7.0837,   50.9140,   -3.1885,   26.1288,   -6.1299,    11.2175,   35.4476,  -30.4055,  -36.4612,
                 -62.0732,   -8.4399,   14.4819,  -51.4732,    1.6699,   77.5096,  -18.6569,    50.1675,   62.6634,   22.5775,   16.9881,
                  90.7339,    8.8552,  -50.9818,   20.9035,    0.7712,  -75.8221,  -19.0744,   -35.3043,   55.4823,  -14.1388,   70.1972,
                 -16.3458,   38.5544,  -25.9921,   17.2767,   49.0289,  -67.9383,   35.1534,    21.0745,   23.8821,   -9.0171,  159.3690,
                  73.7248,  169.3710, -111.7520,  145.2790, -140.8670, -161.6070,  -58.5454,    17.0275,  -88.7363,   -8.1599,   84.4661,
                 -11.5182,  112.5230,   42.2946,  141.6010, -141.0130,  112.6920, -146.3120,   117.7950, -156.5060,  -62.0114, -165.3900,
                 -41.5667,   14.8274, -112.8120,   28.4051,  -66.2272,   18.5640,  110.1050,     4.5585,   13.1075,  172.1850,   51.2246,
                -163.9080,   88.4345, -179.2690, -104.2380,  150.2140,  -53.4349,  177.6410,  -166.9130,  -19.3140, -137.7110,  -21.1100,
                  52.3250,  -50.1040,  130.8480,  -28.1110,   55.9000,  144.9420,   37.0490 \} } \\
  F89 & \texttt{ \{        2.2272,   83.8283,  -17.2516,   62.0451,   -4.3957,   -6.2564,   -66.8426,    2.2260,   -7.5549,    8.0116,
                  19.7190,   45.8951,   59.1790,  -44.2227,  -38.1666,   61.8643,  -12.3786,    66.5324,  -11.5333,   21.1678,   55.8142,
                  -6.0654,   33.8288,   27.3280,    4.9158,    9.6687,    6.6218,  -29.0447,    37.0295,  -69.7487,   55.7643,    5.2775,
                 -85.3894,   19.8100,  -56.6216,   35.6304,   12.7408,  -27.1700,   46.5212,    35.6991,    0.1863,   -1.8802,  -35.9248,
                 -22.4080,   22.5394,   38.3030,   17.4329,  -50.3094,   16.5954,  -13.2596,   -75.6344,    3.0747,   -5.4558,  -32.5393,
                  -2.4337,  -38.7415,    9.2880,    3.5002,  -94.3724,   -7.7578,    9.8094,    47.4793,  -24.8944,   27.0513,    8.0774,
                 -22.1472,  -36.5275,  -28.8212,   19.7025,   81.2552,  -16.7670,   23.8761,   -12.0107,  -25.4997,    5.1184,   14.7353,
                  39.6318,   35.9480,   -8.0517,  -41.7183,   22.3815,    1.0366,   -3.9487,  -149.7020,  -72.6194,   26.2519,   16.0132,
                 143.1950,   22.3287,  160.6860, -144.2450,  109.1530, -146.4920,  130.2590,  -146.3270,  -58.0928, -163.3600, -172.2370,
                 160.7630,  174.8980,   62.5863,   -5.7606,   57.2637,  167.9240,  -11.5373,  -130.6350,  -16.0886,   14.2913,  -55.8666,
                -153.8890,  -61.0903,   22.9111,  -82.1330,   29.6814,  -37.7871,  157.3670,   117.8590,    2.1448,   33.3896, -170.4780,
                  49.2151, -149.0470,  107.5210, -170.4170,  -48.7215, -148.5490,  -25.9835,  -130.2780,  -43.5506,   47.3040,  -44.3866,
                  40.9219,  131.1740,   -2.8724,  160.3930,   47.5093,  -19.5739, -160.8470,   -59.9741, -175.7010, -130.4430,  147.7360,
                  52.9853,  160.5270,  -18.4373,    6.4961,  110.2900,   -5.9111,  146.7160,   140.5410, -167.5110,  -54.9358, -134.6400,
                  135.564, -134.7360,    1.0748,  -50.1584,   28.6888,  121.6580,   17.2979,   137.3170,   28.9442,  112.3580, -153.7090,
                  116.875, -160.8470,  -59.4520, -162.7970,  -49.5730,  -15.6227,   21.0721,    27.9867,  135.005 \} } \\

     \end{tabular}
     }
\end{table*}

\bibliographystyle{spmpsci}


\end{document}